\documentclass{article}
\usepackage{microtype}
\usepackage{multirow}
\usepackage{graphicx}
\usepackage{subfigure}
\usepackage{booktabs} % for professional tables
\usepackage{float}

\usepackage{hyperref}

\usepackage[accepted]{icml2025_arxiv}

\usepackage{amsmath}
\usepackage{amssymb}
\usepackage{mathtools}
\usepackage{amsthm}

\usepackage[capitalize,noabbrev]{cleveref}

\theoremstyle{plain}
\newtheorem{theorem}{Theorem}[section]

\theoremstyle{definition}
\newtheorem{definition}[theorem]{Definition}

\theoremstyle{remark}

\usepackage[textsize=tiny]{todonotes}

\icmltitlerunning{SUPRA: Subspace Parameterized Attention for Neural Operator on General Domains}

\begin{document}

\twocolumn[
\icmltitle{SUPRA: Subspace Parameterized Attention for Neural Operator on General Domains}



\icmlsetsymbol{equal}{*}

\begin{icmlauthorlist}
\icmlauthor{Zherui Yang}{ustc-math}
\icmlauthor{Zhengyang Xue}{ustc-math}
\icmlauthor{Ligang Liu}{ustc-math}
\end{icmlauthorlist}

\icmlaffiliation{ustc-math}{School of Mathematical Sciences, University of Science and Technology of China, Hefei, Anhui, China}

\icmlcorrespondingauthor{Ligang Liu}{lgliu@ustc.edu.edu}


% TODO: Check this.
\icmlkeywords{Machine Learning, PDE, Numerical Analysis}

\vskip 0.3in
]



\printAffiliationsAndNotice{}  % leave blank if no need to mention equal contribution

% ==> Sentence 1
% Neural operators are efficient surrogate models for solving partial differential equations (PDEs), but their key components face challenges: 
%   (1) in order to improve accuracy, attention mechanisms suffer from computational inefficiency on large-scale meshes, and
%   (2) spectral convolutions rely on the Fast Fourier Transform (FFT) on regular grids, which cause accuracy degradation on irregular domains.
% <==== Sentence 1

% ==> Sentence 2: Function Attention
% To tackle these problems, we regard the matrix-vector operations in standard attention mechanism on vectors in Euclidean space as bilinear forms and linear operators in vector spaces, and generalize the attention mechanism to function spaces.

% Q1. Should we emphasize our method is fully equivalent to original one?
% <==== Sentence 2

% ==> Sentence 3: From Function Space to Subspace
% Because of the infinite dimensionality of function spaces, 
% <==== Sentence 3

% ==> Sentence 4: Subspace Parameterization

% <==== Sentence 4

% ==> Sentence 5: Laplacian Eigensubspace
% For irregular domains, we propose the Laplacian eigensubspace in SUPRA.

% <==== Sentence 5

% ==> Sentence 6: Experiments.

% <==== Sentence 6

\begin{abstract}
% ==> Sentence 1
Neural operators are efficient surrogate models for solving partial differential equations (PDEs), but their key components face challenges: 
  (1) in order to improve accuracy, attention mechanisms suffer from computational inefficiency on large-scale meshes, and
  (2) spectral convolutions rely on the Fast Fourier Transform (FFT) on regular grids and assume a flat geometry, which causes accuracy degradation on irregular domains.
% ==> Sentence 2: Function Attention
To tackle these problems, we regard the matrix-vector operations in the standard attention mechanism on vectors in Euclidean space as bilinear forms and linear operators in vector spaces and generalize the attention mechanism to function spaces.
% ==> Sentence 3: From Function Space to Subspace
This new attention mechanism is fully equivalent to the standard attention but impossible to compute due to the infinite dimensionality of function spaces.
% ==> Sentence 4: Subspace Parameterization
To address this, inspired by model reduction techniques, we propose Subspace Parameterized Attention (SUPRA) neural operator, which approximates the attention mechanism within a finite-dimensional subspace.
% ==> Sentence 5: Laplacian Eigensubspace
To construct a subspace on irregular domains for SUPRA, we propose using the Laplacian eigenfunctions, which naturally adapt to domains' geometries and guarantee the optimal approximation for smooth functions.
% ==> Sentence 6: Experiments.
Experiments show that the SUPRA neural operator reduces error rates by up to 33\% on various PDE datasets while maintaining state-of-the-art computational efficiency.

\end{abstract}
% Neural operators have emerged as efficient surrogate models to classical solvers for solving partial differential equations (PDEs).
% However, their key components still face challenges: attention mechanisms struggle to balance accuracy and computational efficiency on large-scale meshes, while spectral convolutions rely on the Fast Fourier Transform (FFT) on regular grids, which can be unsuitable for irregular domains.
% To address these issues, we introduce the Subspace Parameterized Attention (SUPRA) neural operator, a novel method that directly extends the standard attention mechanism from finite-dimensional vector spaces to infinite-dimensional function spaces.
% To enable efficient computation, SUPRA parameterizes attention within a finite-dimensional subspace, achieving a balance between performance and computational cost.
% For irregular domains, we leverage the Laplacian eigenfunctions as the subspace basis, which naturally adapt to the domains' geometry, enabling more accurate approximations.
% Experiments show that the SUPRA neural operator reduces error rates by up to 33\% on various PDE datasets while maintaining state-of-the-art computational efficiency.

% including finite difference methods \cite{hesthaven2018numericalmethod}, and spectral methods \cite{bernardi1997spectral},
\section{Introduction}
Partial Differential Equations (PDEs) are critical in modeling physical and engineering systems, such as weather forecasting \cite{bonev2023sphericalfn}, fluid dynamics \cite{horie2024conservation}, and structural analysis \cite{li2022geofno}.
Solving PDEs has historically relied on numerical methods, such as finite element methods \cite{brenner2008fem}, while they often become computationally expensive, especially for large-scale meshes and irregular domains \cite{domaindecomposition}. 
Recently, deep learning methods have shown great potential in accelerating PDE solving \cite{karniadakis2021physicsinformedml}. By learning mappings from inputs to solutions, neural operators \cite{kovachki2023neuraloperator}, such as Fourier Neural Operators (FNOs) \cite{li2020fourierno}, and DeepONet \cite{lu2019learningno}, offer significant speed-ups compared to traditional methods, while also being flexible and scalable.

\begin{figure}[htbp]
    \centering
    \includegraphics[width=1\linewidth]{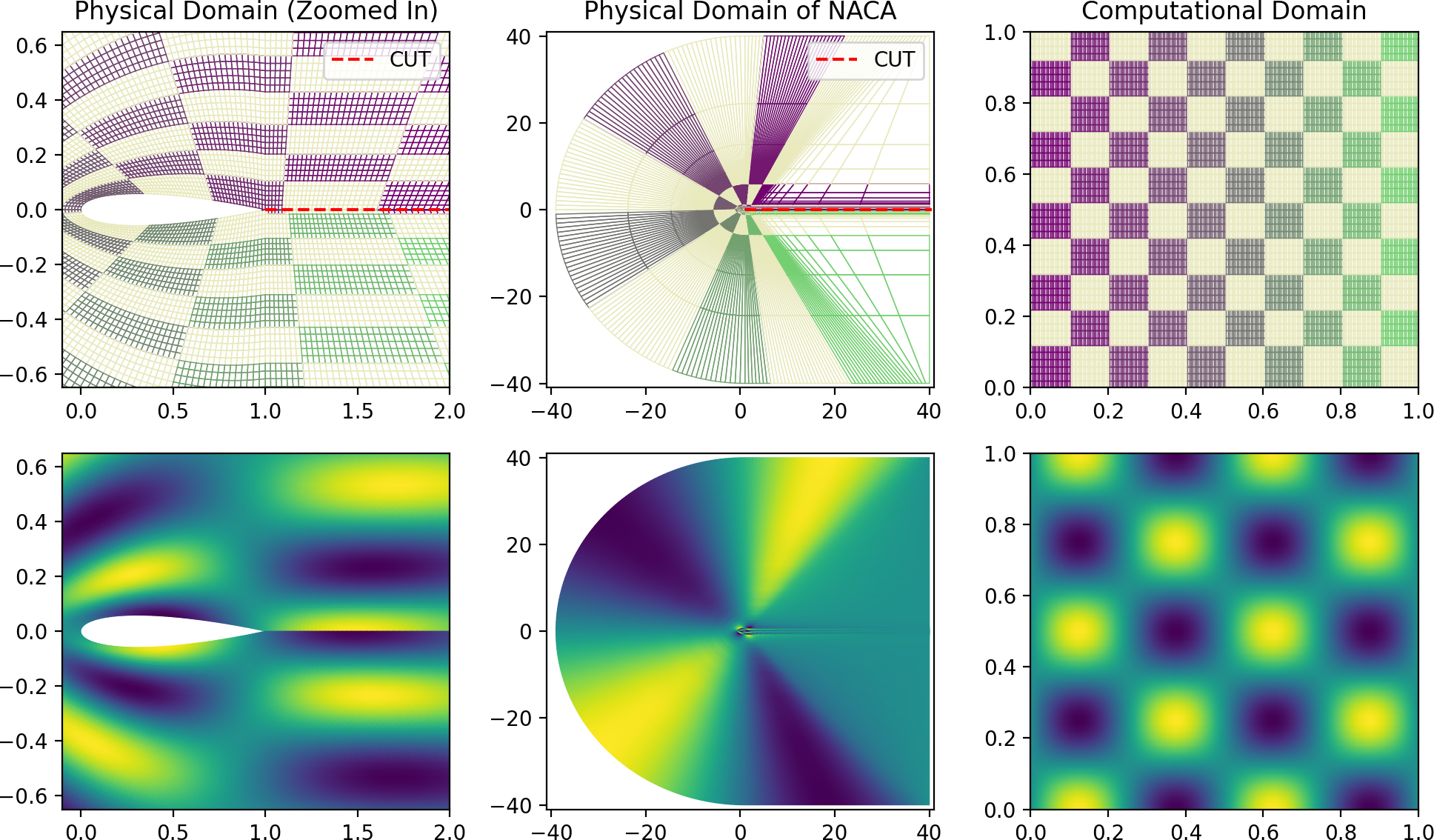}
    \caption{Discontinuities induced by domain cuts. The top row shows the mapping between a physical domain and a computation domain in the NACA Airfoil problem. 
    To define the continuous map, the physical domain must be cut along the homology loop (red dashed line).
    Figures in the bottom row visualize a continuous function $f(x, y) = \sin(4\pi x) \cos(4 \pi y)$ in different domains, where $x, y$ are coordinates in the computational domains. After mapping back to the physical domain, the continuity of the function does not hold because of the cut.}
    \label{fig:naca-intro}
\end{figure}

Despite these advances, neural operators still face deficiencies in addressing the challenges of complex physical fields (e.g. Navier-Stokes equations with high Reynolds number) and irregular computational domains in multi-scale problems (e.g. NACA Airfoil). 
There are two main problems with existing approaches:
(1) \textbf{The struggle between computational complexity and expressive power of attention}. As one of the most fundamental components of deep learning, the attention mechanism \cite{vaswani2017attentionia} has also been introduced in PDE-solving tasks. Fourier attention \cite{cao2021chooseat} treats sample points as tokens, resulting in a time complexity that is quadratic in the number of sample points, which is computationally expensive for large-scale meshes. Galerkin attention reduces the complexity to linear levels, forcing dot products to be applied only in the space spanned by the input functions \cite{wang2024lno} and limiting the expressive power of the attention mechanism. Although additional modules, such as FNOs \cite{rahman2024pretraining}, can be used to improve accuracy, they also introduce additional overhead and constraints.
(2) \textbf{Discontinuities in functions induced by cuts to irregular domains.} (see Figure \ref{fig:naca-intro}). In order that FFT can be applied to conduct spectral convolution \cite{li2020fourierno}, irregular physical domains are cut and mapped to a regular computational grid \cite{li2022geofno}. 
However, the cut introduces discontinuities to smooth functions defined on the computational grid after they are mapped back to the original physical domain. 
The subsequent works extend the FFT to spheres \cite{bonev2023sphericalfn} or point clouds \cite{wang2024beno}, but there still lacks a discussion on constructing suitable basis functions for more general domains and complex geometries.

% Functions are the fundamental elements in the context of neural operator and PDE solving, analogous to tokens in NLP tasks. Based on this idea, to address the challenges above, we derive the attention mechanism on function spaces by regarding the matrix-vector operations as bilinear forms and linear operators on vector spaces. 
\newpage
To address the challenges above, we consider functions as fundamental elements in neural operators and PDE solving tasks, analogues to tokens in NLP tasks.
Based on this idea, we derive the attention mechanism on function spaces by regarding the matrix-vector operations as bilinear forms and linear operators on vector spaces. 
This attention mechanism is fully equivalent to the standard defined on $\mathbb{R}^N$, aiming to capture complex relations between input functions. 
To tackle the inefficiency of computation in infinite-dimensional function spaces, we draw inspiration from model reduction techniques and propose Subspace Parameterized Attention (SUPRA), which parameterizes the attention mechanism within a finite-dimensional subspace. SUPRA transforms function-wise attention into standard attention operations on subspace coordinates, balancing its expressive power and computational efficiency.

To construct a suitable subspace for SUPRA on irregular domains, we leverage the subspace using the eigenfunctions of the Laplace operator. These eigenfunctions are obtained by solving the Laplacian eigenvalue problem defined by the geometry. They naturally encode the topology of the physical domain, ensure continuity across irregular meshes, and have excellent approximation properties for smooth functions. Leveraging their continuity and approximation properties, the SUPRA neural operator accurately captures the relations between functions while maintaining low computational costs on irregular domains.

In summary, our contributions are as follows:
\begin{enumerate}
    \item We formulate the attention mechanism for infinite-dimensional function spaces and parameterize the attention between functions within finite-dimensional subspace. To the best of our knowledge, our method is the first to directly extend the attention to function spaces while achieving a balance between attention's expressive power and computational efficiency.
    \item We leverage the Laplacian eigenfunctions to construct the subspace for irregular domains, which guarantees both the continuity of basis functions and optimal approximation property for smooth functions.
    % \item We present a Laplacian-based view for selecting and constructing subspace basis functions. This approach ensures the continuity of basis functions in irregular physical domains.
    \item Through experiments and ablation studies on various PDE datasets, we demonstrate that our method achieves superior accuracy and computational efficiency compared to existing approaches.
\end{enumerate}

\section{Related Works}\label{sec:related-works}

\subsection{PDE Solvers}

Solving PDEs is critical in many scientific research and industrial applications. Classical solvers, such as finite difference methods \cite{hesthaven2018numericalmethod},  finite element methods \cite{brenner2008fem}, and spectral methods \cite{bernardi1997spectral}, typically discretize the physical domain and solve the resulting linear systems. While well-established, these approaches are computationally expensive on high-dimensional problems or fine grids, driving interest in alternative solutions \cite{domaindecomposition}.

\textbf{Physics-Informed Neural Networks (PINNs)} Through defining a loss using PDEs' residuals, PINNs use neural networks to represent the solution function directly, taking spatiotemporal coordinates as inputs and force the neural network to predict solution values under given initial and boundary conditions directly \cite{raissi2017physicsi, raissi2019physics}. However, training a PINN typically addresses only one set of initial/boundary conditions and external forces and is not cheap even on modern hardware \cite{rathore2024pinnchallenges}.

\textbf{Neural PDE Solvers} Neural operators learn to approximate the mapping from inputs to solutions \cite{lu2019learningno}. Serving as surrogate models for numerical solvers. This approach significantly reduces the computational cost of solving PDEs approximately. As one of the pioneering works, FNO \cite{li2020fourierno} and its variants \cite{kossaifi2023multigridtf, wen2022ufno} have achieved a significant leap in solving PDEs. FNOs use the Fast Fourier Transform (FFT) to perform efficient spectral convolution on regular grids. Although subsequent works, such as SFNO \cite{bonev2023sphericalfn}, have extended the transformations to spherical domains using spherical harmonic transforms, studies on general regions remain limited. Geo-FNO \cite{li2022geofno} utilize mappings to transform irregular domains into regular rectangular regions, while the existence of continuous mappings is not guaranteed. GINO \cite{li2023geometryinformedno} propagates information between irregular and regular domains through a GNO, but it does not fundamentally solve the problem of spectral convolution on irregular grids. Although Fourier transform can be extended to point clouds \cite{lingsch2024beyoundregulargrids}, the Fourier basis is also unsuitable for irregular domains and multi-scale problems.

\subsection{Attention Mechanism in Neural PDE Solvers}

As a breakthrough innovation in deep learning, the attention mechanism and Transformer model \cite{vaswani2017attentionia} have also been extensively applied to solving PDEs. In Fourier attention \cite{cao2021chooseat}, point features are regarded as tokens, resulting in an attention weight matrix that scales quadratically with the number of points. Many works, such as Fact-Former\cite{li2023factformer}, aim to reduce this computational complexity but are limited to regular domains. In contrast, others \cite{li2022oformer, xiao2023improvedol, hao2023gnot} leverage linear transformers to address quadratic complexity by reordering computations but restricting attention to the subspace spanned by input functions. CoDA-NO\cite{rahman2024pretraining} use FNOs instead of linear combinations to generate attention inputs. Transolver \cite{wu2024transolveraf} extracts global features by dividing the input physical field into slices to enable global information exchange.  LNO \cite{wang2024lno} aggregate point data into global features to apply attention while failing to achieve high accuracy on irregular domains.

\subsection{Model Reduction}

Model reduction reduces computational complexity by projecting high-dimensional state spaces into low-dimensional subspaces, preserving key properties of the original system \cite{mor2017survey}. Model reduction has many real-world applications. Proper Orthogonal Decomposition methods (POD) \cite{sirovich1987turbulenceat} and Principle Component Analysis (PCA) \cite{brunton2019machinelf} are standard model reduction techniques in fluid dynamics to transform from physical coordinates into a modal basis. In structural mechanics problems, model reduction also helps to accelerate the simulation \cite{sifakis2012femsimulation3d} or optimization problems \cite{choi2020gradientbased, siam2015projmodelreduction}. 

%Subspace methods can also be applied in many optimization problems \cite{liu2021subspaceoptimization}.

\section{Preliminaries}\label{sec:preliminaries}

\subsection{Problem Setup}
As an instance, we consider a PDE with a boundary condition defined on domain $\Omega\subseteq \mathbb{R}^d$,
\begin{equation}
\begin{aligned}
    \mathcal{L}_{a} u & = 0, &\quad & x \in \Omega,\\
    u &= 0, &\quad & x \in \partial \Omega,
\end{aligned}
\end{equation}
where $\mathcal{L}_{a}$ is an operator that composed of the given function $a(x)$ and partial differentials of the unknown function $u(x)$.

The role of neural operators is to define mappings between functions. In the aforementioned problem, neural operators learn the mapping from the parameter to the solution, i.e. $a\mapsto u$. 
To enable computation on functions, both the input and output functions are represented by their values at $M$ sample points $x_i$ $(1 \le i \le M)$.

\textbf{Motivation of Generalizing Attention to Functions} In the context of neural operators, functions are naturally treated as the fundamental "primitives". Generalizing the attention mechanism to function spaces losslessly is imperative, as it allows us to model complex relations between functions. In NLP tasks, attention is applied to tokens, represented as vectors in finite-dimensional space $\mathbb{R}^N$. To generalize the attention mechanism from vectors to functions, we first introduce two key concepts in this paper.

\subsection{Linear Operators and Bilinear forms}

\begin{definition}[Linear Operator $b(\cdot)$]
Given a vector space $\mathbf{V}$, a linear operator is a mapping $b(\cdot): \mathbf{V} \to \mathbf{V}$ if it satisfies both additivity and homogeneity \cite{stein2009real}
\begin{equation}
    \forall u, v \in \mathbf{V}, \alpha \in \mathbb{R}, \quad b(\alpha u + v) = \alpha b(u) + b(v).
\end{equation}
\end{definition}
    
\begin{definition}[Bilinear Form $a(\cdot, \cdot)$]
    A bilinear form is a mapping $a(\cdot, \cdot): \mathbf{V}\times \mathbf{V} \to \mathbb{R}$ if for all $u, v, w \in V$ and $\alpha \in \mathbb{R}$
    \begin{equation}
        A(\alpha u + v, w) = \alpha A(u, w) + A(v, w),
    \end{equation}
    and similarly for its second argument.
\end{definition}

In the rest of this paper, we will focus linear operators and bilinear forms on $\mathbf{V}=\mathbb{R}^N$ or $\mathbf{V}=L^2(\Omega)$ since they are used to define the weights and outputs of attention mechanisms.

% \textbf{Linear Operator and Bilinear Forms} Given a vector space $\mathbf{V}$ (e.g. Euclidean space $\mathbb{R}^N$, function space $L^2(\Omega)$) and a field $K$ (e.g. $\mathbb{R}$, $\mathbb{C}$), a map $b(\cdot): \mathbf{V} \to \mathbf{V}$ is called a linear operator if it satisfies both additivity and homogeneity: %of degree 1:
% \begin{equation}
% \begin{aligned}
%     \forall u, v, \in \mathbf{V}, \quad& b(u +v) =  b(u) + b(v), \\
%     \forall \alpha\in K,  \quad& b(\alpha u) = \alpha b(u), 
% \end{aligned}
% \end{equation}
% and a bilinear form $a(\cdot, \cdot)$ is map $\mathbf{V}\times \mathbf{V} \to K$ that is linear in each argument separately:
% \begin{equation}
% \begin{aligned}
%     \forall u, a(u, v) \text{ is a linear operator},\\
%     \forall v, a(u, v) \text{ is a linear operator}.
% \end{aligned}
% \end{equation}
% In the rest of this paper, we will focus on $K=\mathbb{R}$ and $\mathbf{V}=\mathbb{R}^N$ or $\mathbf{V}=L^2(\Omega)$ since they are used to define the weights and outputs of general attention mechanism. 

\subsection{Standard Attention}
Self-attention weights between two tokens $x_i, x_j\in \mathbb{R}^N$, $1 \le i , j \le C$ are defined with their query vectors $q_i, q_j\in  \mathbb{R}^N$ and key vectors $k_i, k_j\in \mathbb{R}^N$ as follows \cite{vaswani2017attentionia}:
\begin{equation}
    w_{ij} = q_i^\top k_j, \quad 1 \le i,j \le C,
\end{equation}
where $q_i, k_i, v_i$ are linear transform of $x_i$:
\begin{equation}\label{eq:qkv-definition}
    q_i = W_Q x_i, \quad k_i = W_K x_i, \quad v_i = W_V x_i.
\end{equation}

In \cref{eq:qkv-definition} $W_Q, W_K, W_V \in \mathbb{R}^{N \times N}$ are learnable matrices. The attention weights before softmax operation can be formulated as
\begin{equation}\label{eq:attn-weight-token}
  w_{ij} = \frac{1}{\sqrt{N}} x_i^\top W_Q^\top W_K x_j.
\end{equation}

Finally, the outputs of attention $z_i$ are defined as
\begin{equation}\label{eq:attn-output}
  z_i = \sum_{j=1}^C \frac{\mathrm{exp}(w_{ij})}{\sum_{k=1}^C \mathrm{exp}(w_{ik})}v_j.
\end{equation}

\section{Our Method}\label{sec:method}
% In this section, we first extend the definition of the attention mechanism to general vector spaces, particularly infinite-dimensional functional spaces. Then, we demonstrate how to parameterize the attention in a finite-dimensional subspace and the overall design of our neural operator. We also show our Laplacian-based view for constructing the subspace.

\subsection{Attention Mechanisms on Function Spaces}

% \textbf{Attention on General Vector Spaces} 

As shown in \cref{eq:attn-weight-token}, attention weights are essentially computed from a bilinear form $w_{ij} = a(x_i, x_j) = x_i^T W_Q^T W_K x_j$ in the vector space $\mathbf{V} = \mathbb{R}^N$, while in \cref{eq:qkv-definition}, $v_i$ are computed from a linear operator $v_i = b(x_i) = W_V x_i$. In this case, the attention operation's outputs $z_i$ with input $x_i \in \mathbb{R}^N$ are
\begin{equation}\label{eq:general-attn-output}
    z_i = \sum_{j=1}^N \frac{\exp{a(x_i, x_j)}}{\sum_{k=1}^N \exp{a(x_i, x_k)} } b(x_j).
\end{equation}

To define attention on any vector space, only a bilinear form $a(\cdot, \cdot)$ and a linear operator $b(\cdot)$ are needed. 
In the rest of the paper, we will focus mainly on the function space $\mathbf{V} = L^2(\Omega)$.

\textbf{Attention on Function Spaces} Let $u_i, u_j \in L^2(\Omega), 1\le i, j \le C$ be functions defined on domain $\Omega$, suppose $\mathcal{B} = \{e_k\}$ is a basis of $L^2(\Omega)$, $u_i$ can be represented by the basis by $u_i = \sum_k \hat{u}_i^k e_k$.
The coefficients $\hat{u}^k$ represent the coordinates of the function $u$ under the basis. By substituting $u_i, u_j$ into the bilinear form $a(u_i, u_j)$ and the linear operator $b(u_i)$, and using their linear properties, the bilinear form and linear operator are represented by the basis $\mathcal{B}$ as
\begin{equation}\label{eq:attn-inf-space}
\begin{aligned}
a(u_i, u_j) &= a\left(\sum_k \hat{u}_i^k e_k,\sum_l \hat{u}_j^l e_l\right) = \sum_{k, l} a(e_k, e_l) \hat{u}_i^k \hat{u}_j^l, \\
 b(u_i) &= \sum_{k}\hat{b}^k(u_i) e_k = \sum_{k, l} \hat{b}^k(e_l) {\hat{u}_i^l}  e_k.
\end{aligned}
\end{equation}

In \cref{eq:attn-inf-space}, $a(e_k, e_l), \hat{b}^k(e_l)\in\mathbb{R}$ are learnable coefficients, that do not depend on the input functions $u_i$, but only on the basis. \cref{eq:general-attn-output} and \cref{eq:attn-inf-space} are direct generalizations of the standard attention. However, since the function space $L^2(\Omega)$ is infinite-dimensional, infinitely many coefficients are impossible to compute.

\subsection{Efficient Attention on Function Spaces}

\textbf{Subspace Parameterized Attention} Inspired by model reduction techniques, the bilinear form can be efficiently approximated by truncating the basis at the $N$-th term if the basis $\mathcal{B}$ has a good approximation property:
\begin{equation}\label{eq:attn-parameterized}
\begin{aligned}
a(u_i, u_j)
& \approx \sum_{k, l = 1}^N a(e_k, e_l) \hat{u}_i^k \hat{u}_j^l
= \hat{u}_i^\top A\hat{u}_j, \\
% \hat{b}^k(u_i) & \approx \sum_l \hat{b}^k(e_l) {\hat{u}_i^l} e_k = B_{k, :} \hat{u}_i
b(u_i) & \approx \sum_{k, l=1}^N\hat{b}^k(e_l) {\hat{u}_i^l} e_k, \text{ and } \hat{b}(u_i) \approx B \hat{u}_i.
\end{aligned}
\end{equation}
In \cref{eq:attn-parameterized}, matrix $A = [a(e_k, e_l)]_{N\times N}$ parameterizes the bilinear form, $B =[\hat{b}^k(e_l)  ]_{N\times N}$ parameterizes the linear operator within the subspace, and $\hat{u}_i \in \mathbb{R}^N$ are vectors composed of the coefficients $\hat{u}_i^k$. 
Plugging the parameterizations into the \cref{eq:general-attn-output}, the final output functions $z_i, 1\le i \le C$ of attention are
\begin{equation}\label{eq:supra-main}
\begin{aligned}
    z_i= \sum_{l=1}^N \hat{z}_i^l e_l, \text{ where } \hat{z}_i= \sum_{k=1}^C \frac{\mathrm{exp}(w_{ik})}{\sum_{k=1}^C \mathrm{exp}(w_{ik})} B \hat{u}_k,
\end{aligned}
\end{equation}

At this point, the attention mechanism on the function space has been defined (in \cref{eq:general-attn-output} and \eqref{eq:attn-inf-space}) and parameterized within a subspace (in \cref{eq:attn-parameterized} and \eqref{eq:supra-main}). Therefore, we refer to this attention mechanism as \textbf{Subspace Parameterized Attention (SUPRA)}. As proved in \cref{subsec:approx-bilinear-bounded}, as $N$ increases, SUPRA can approximate the attention in function spaces with arbitrary accuracy.

\textbf{Subspace Projection and Reconstruction} 
Besides operations on coordinates, projection from function $u$ to subspace coordinates $\hat{u}$ and reconstruction from $\hat{u}$ to $u$ are also required in SUPRA. 
Reconstruction is a trivial linear combination of the basis functions, while projection involves numerical integration. For common cases, such as regular grids $[0,1]^2$ with resolution $H\times W$, the projection onto an orthonormal basis is simply a weighted sum,
\begin{equation}
\hat{u}^k = \int_{[0, 1]^2} u(x) e_k(x) \mathrm dx \approx \frac{1}{H W} \sum_{i, j=1}^{H, W} u(x_{ij}) e_k(x_{ij}),
\end{equation}
where $x_{ij} = (i / H, j / W)$. This operation can also be conducted efficiently on GPUs.

\begin{figure*}[t]
        \centering
        \includegraphics[width=1.0\linewidth]{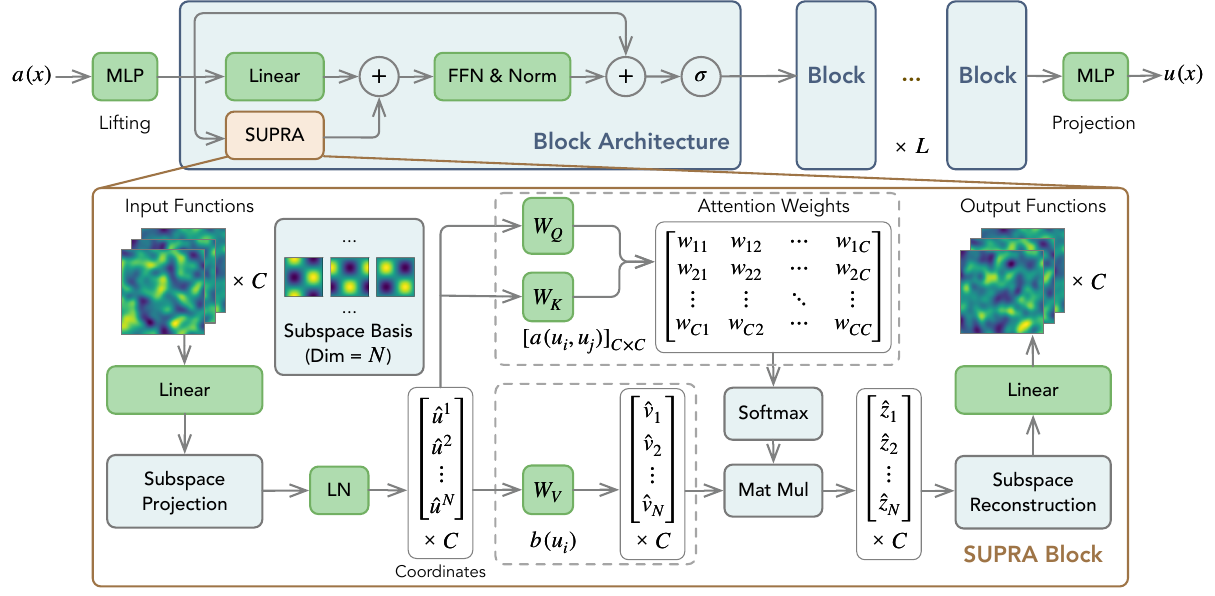}
        \caption{Overall design of SUPRA neural operator. We adopt the architecture proposed in \cite{kossaifi2023multigridtf} while replacing spectral convolutions with SUPRA blocks. All trainable modules are colored green. LN stands for LayerNorm, $W_V$ corresponds to the matrix $B$, and $W_Q^\top W_K$ corresponds to the matrix $A$ defined in \cref{eq:attn-parameterized}. Although LayerNorm \cite{ba2016layern} is a common choice, InstanceNorm \cite{ul2016instancenorm} can work better since each function is treated as a token in our framework.}
        \label{fig:pipeline}
\end{figure*}

\subsection{Subspace Construction for General Domains}\label{subsec:basis}

The subspace spanned by basis functions serves as the arena for the attention mechanism, influencing how functions engage and exchange information. On regular domains, any orthonormal basis with promising approximation properties can work in SUPRA, such as Chebyshev \cite{chebyshev_basis} and Fourier basis \cite{fourier_basis}. For irregular domains, we provide a Laplacian-based method to construct proper basis functions \cite{spherical_harmonics_basis}.

\textbf{Laplacian Eigensubspace} The Laplacian eigensubspace is spanned by the smallest eigenfunctions of the Laplace operator, which are the Fourier basis on regular grids and spherical harmonics on spheres. \textit{Therefore, the Laplacian eigensubspace can be interpreted as a natural extension of the Fourier basis from regular grids to general domains}.

\begin{figure}[t]
    \centering
    \includegraphics[width=0.85\linewidth]{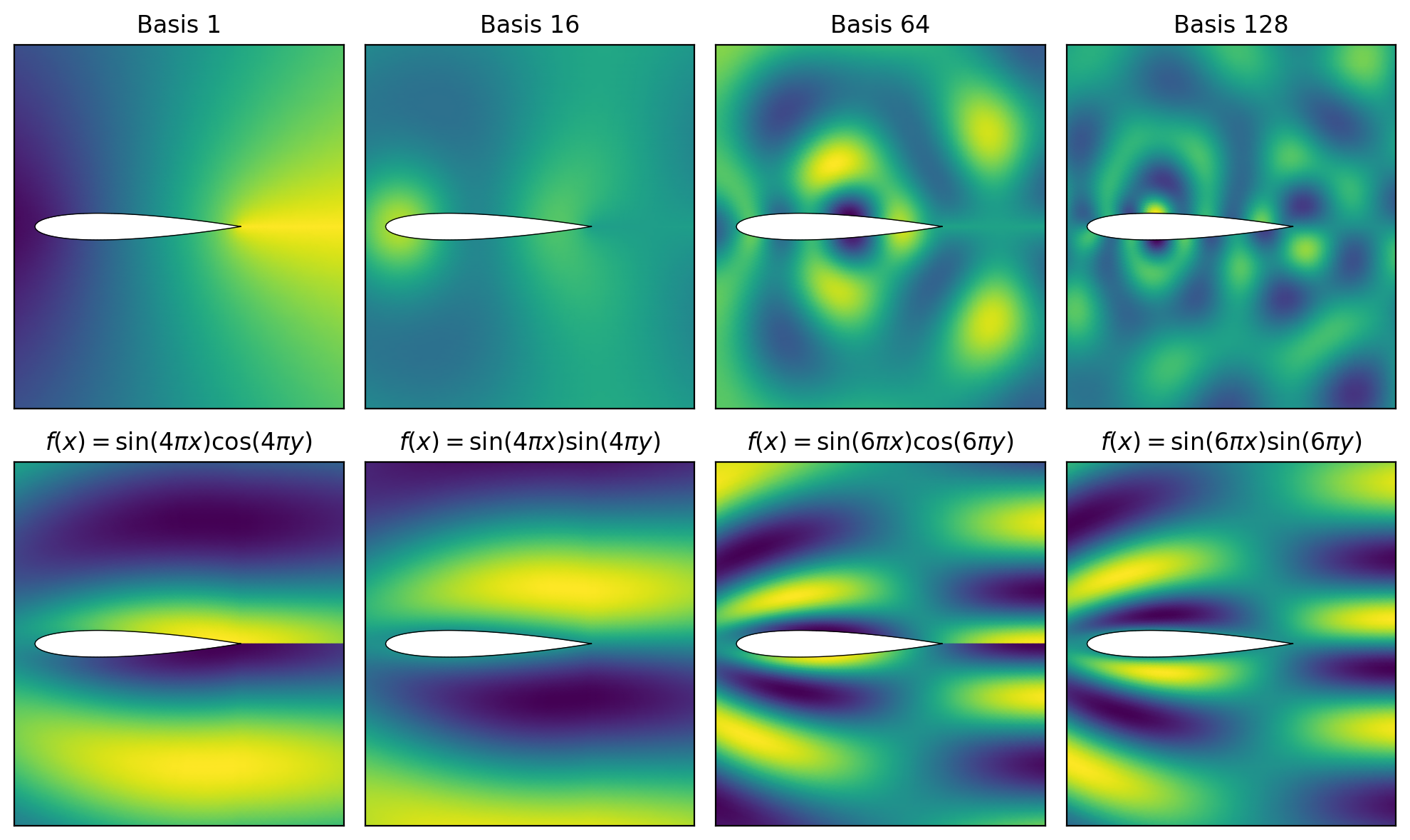}
    \caption{Comparison between Laplacian eigenfunctions and Fourier basis. The eigenfunctions guarantee continuity across the physical domain, while the Fourier basis defined on the computational mesh does not.}
    \label{fig:naca-subspace}
\end{figure}

The Laplacian eigensubspace guarantees optimal approximation for smooth functions that adapt to the geometry of any given domain. These eigenfunctions are naturally orthonormal and continuous (see \cref{fig:naca-subspace}), ensuring efficient dimensionality reduction while reflecting the domain's topology (see also \cref{sec:topo-change}).

\textbf{Basis Construction}  For irregular domains, the eigenvectors of the Laplace matrix resulting from FEM discretization are precomputed once by using classical methods. For regular grids, the basis is computed from a tensor product. See \cref{subsec:abl-study} and \cref{sec:experiments-full} for their implementation details. Typically, $N=64$ to $N=256$ is sufficient for SUPRA.

% \subsection{Practical Neural Operator Design}\label{subsec:operator-design}
% As shown in \cref{fig:pipeline}, we follow the optimized architecture of neural operators \cite{kossaifi2023multigridtf} and propose to replace spectral convolution with the SUPRA block to achieve global operation on function spaces.
% \begin{remark}
%     Although LayerNorm \cite{ba2016layern} is a common choice in operator learning, InstanceNorm \cite{ul2016instancenorm} could work better since each function is treated as a token in our framework.
% \end{remark}

\subsection{Further Discussions}

\textbf{Multiple heads in SUPRA} To enhance SUPRA's expressive power, we use multiple attention heads by decomposing $A = W_Q^T W_K$ and applying attention to subvectors within each head. Focusing on the parameterized attention in \cref{eq:attn-parameterized}, it is clear that SUPRA can directly leverage highly optimized multi-head attention code, as shown in the SUPRA block in \cref{fig:pipeline}.

\textbf{Comparison to Standard Attention}
Comparing the matrix $A$ to $W_Q^\top W_K$ \cref{eq:attn-weight-token} and the matrix $B$ to $W_V$ in \cref{eq:qkv-definition}, it is clear that SUPRA transforms the attention between functions $u_i \in L^2(\Omega)$ into standard attention between the coordinate vectors $\hat{u}_i\in\mathbb{R}^N$ of these functions within the subspace.

\textbf{Comparison to Galerkin Attention}
Given $C$ input functions $u_i$, $1\le i \le C$, attention weights in Galerkin attention are defined as the dot product between $v_i, k_i$, $w_{ij} = \langle v_i, k_j \rangle$, and the output is $z_i = \sum_{j=1}^N w_{ij} q_j$. Here, functions $v_i, k_i, q_i$ are \textit{linear combinations} of input functions $u_i$, forcing attention to be applied only within the subspace spanned by $\{u_i\}_{i=1}^C$. 
In contrast, SUPRA is based on bilinear forms and linear operators in a larger function space, allowing it to capture functions' relations and apply transforms directly to each function.

\begin{table*}[t]
    \centering
    \begin{tabular}{c|ccccc|ccc}
\toprule
  \multirow{2}{*}{Test Case} & \multicolumn{5}{c|}{Relative $L^2$ Error ($\times 10^{-2}$)}  & \multicolumn{3}{c}{\#Params ($\times 10^6$)} \\
      & Galerkin & GNOT & Transolver & LNO & Ours & Transolver & LNO & Ours  \\
\midrule
    Darcy & 0.84 & 1.05 & 0.50 & \underline{0.49} & \textbf{0.43}  & 2.8 & \textbf{0.76} & \underline{1.7} \\
    Navier Stokes & 14.0 & 13.8 & \underline{7.83} & 8.45 & \textbf{6.25}  & 11.2 & \underline{5.0} & \textbf{3.4} \\
    Plasticity & 1.20 & 3.36 & \underline{0.08} & 0.29 & \textbf{0.04} & 2.8 & \underline{1.4} & \textbf{1.3} \\
    Airfoil & 1.18 & 0.76 & \underline{0.43} & 0.51 & \textbf{0.34} & 2.8 & \underline{1.4} & \textbf{0.5} \\
    Pipe & 0.98 & 0.47 & 0.32 & \textbf{0.26} & \underline{0.31} & 3.0 & \underline{1.4} & \textbf{1.1} \\
\bottomrule
    \end{tabular}
    \caption{Performance Comparison. Relative $L^2$ error ($\times 10^{-2}$) is recorded, and a smaller value indicates better performance. The best recorded on each dataset is in bold, and the second best is underlined. The best configurations of our method are listed in \cref{subsec:optimal-conf}.}
    \label{tab:std-bench}
\end{table*}

\textbf{Complexity Analysis} SUPRA is efficient compared to all previous methods using the attention mechanism. For $C$ functions sampled at $M$ points in the domain, the complexity of different operations in SUPRA is: (1) projection to subspace coordinates: $O(C M)$, (2) attention between coordinates: $O(C^2 N)$, and (3) reconstruction from coordinates: $O(C M)$. The total complexity of SUPRA is $O(C^2 N + C M)$. 
Correspondingly, Fourier attention considers points as tokens, requiring a complexity of $O(C M^2)$, while Galerkin attention requires a complexity of $O(C^2 M)$. The complexity of SUPRA is similar to Galerkin attention, but achieves better expressive power.

\section{Experiments}\label{sec:experiments}

\textbf{Test Cases} We adopt five standard benchmark datasets provided by the community listed in \cref{tab:test-cases-short}. These test cases cover most of the input types in PDE problems, including changes to the parameters of the equation (Darcy), external inputs (Plasticity), time advance (Navier Stokes), and irregular domains (Pipe and Airfoil). See \cref{sec:experiments-full} for their background information and our experiment's settings.

\begin{table}[h!]
    \centering
    \begin{tabular}{c|cc}
    \toprule
        Test Case & Input & Output \\
    \midrule
        Darcy &  Porus Medium & Fluid Pressure \\ 
        Navier Stokes & Previous Vorticity & Future Vorticity \\
        Plasticity & External Force & Deformation \\
        Airfoil & Structure & Mach Number \\
        Pipe & Structure & Fluid Velocity \\
    \bottomrule
    \end{tabular}
    \caption{List of our test cases and their inputs and outputs.}
    \label{tab:test-cases-short}
\end{table}

\subsection{Accuracy Comparison}

\textbf{Baselines} SUPRA neural operator is compared with the most recent SOTA methods: (1) Galerkin Transformer \cite{cao2021chooseat}, (2) GNOT \cite{hao2023gnot}, (3) Transolver \cite{wu2024transolveraf}, and (4) LNO \cite{wang2024lno}.

\begin{figure}[htbp]
    \centering
    \includegraphics[width=0.95\linewidth]{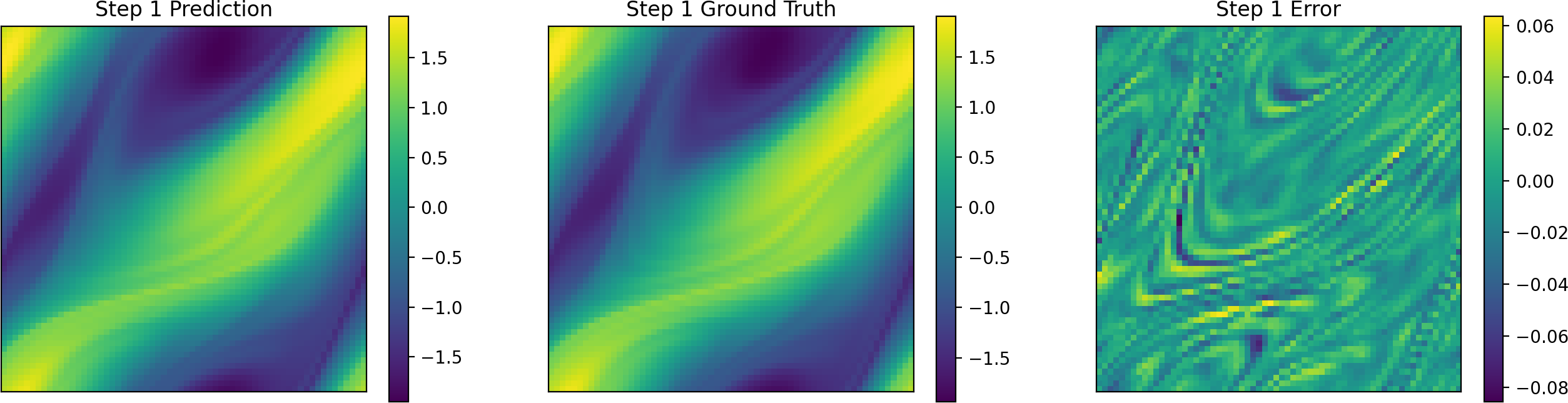}
    \includegraphics[width=0.95\linewidth]{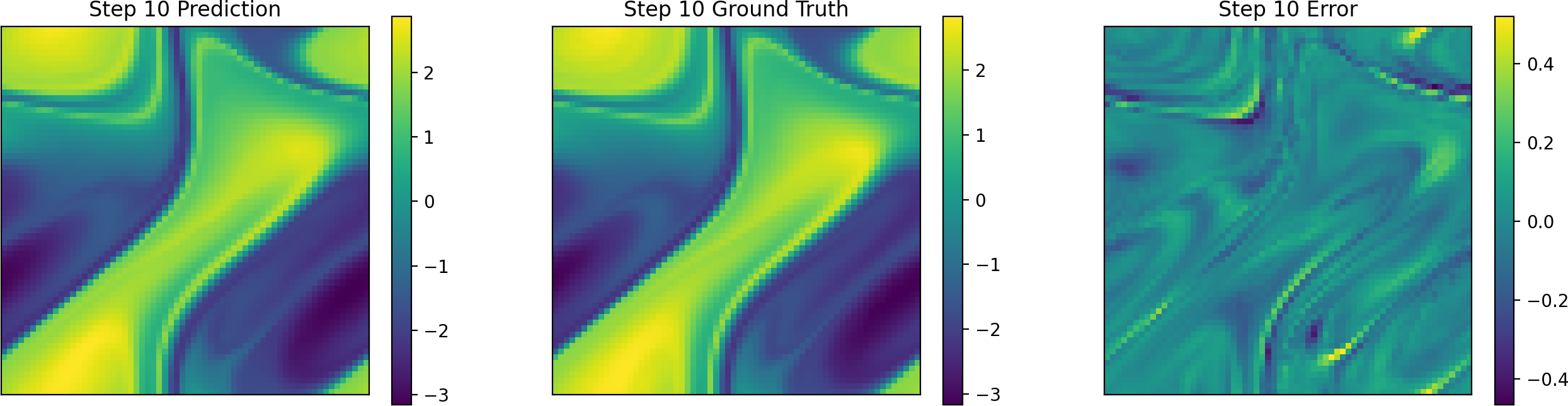}
    \caption{Comparisons between our prediction and ground truth at the first and last step in the Navier Stokes problem. Although the previous input is smooth, the output can be very sharp.}
    \label{fig:ns-compare}
\end{figure}

\textbf{General Settings and Metrics} We use the following relative $L^2$ error (i.e. relative root MSE) as our error metric:
\begin{equation}
    \mathrm{rel}_{L^2}(u, u^*) = \frac{\| u - u^*\|_2}{\|u^*\|_2}, \quad \|f\|_2 = \sqrt{\sum_{i} f(x_i)^2},
\end{equation}
where $x_i$ are sample points, $u$, $u^*$ corresponds to the model prediction and the ground truth.%, and $\| \cdot \|_2$ corresponds to the norm of $L^2$ function space. 
The final relative $L^2$ error is averaged over all test samples. The best performance of each method on different test cases is shown in \cref{tab:std-bench}.

As shown, the SUPRA neural operator achieves the best accuracy and lower computational cost in most test cases. Our model achieves performance close to the SOTA with lower computational cost in the Pipe case.

\begin{figure}[htbp]
    \centering
    \includegraphics[width=1\linewidth]{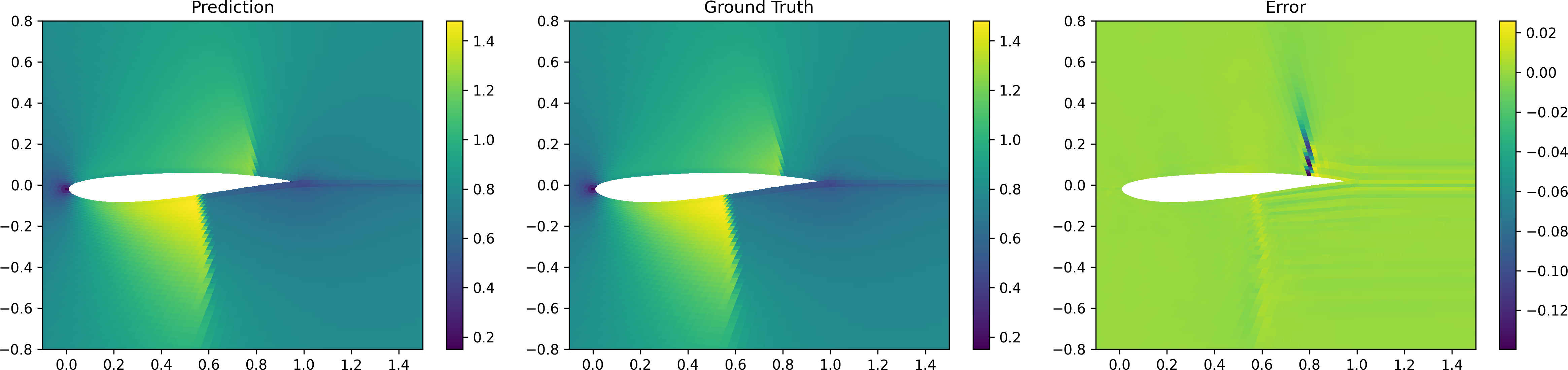}
    \caption{Comparison between our prediction and ground truth in Airfoil problem. Our method can capture the shock wave around the wing precisely.}
    \label{fig:naca-compare}
\end{figure}

We visualize the prediction of the SUPRA neural operator for the Navier Stokes and Airfoil problems in \cref{fig:ns-compare} and \cref{fig:naca-compare}. SUPRA neural operator accurately captures the vortex and shock wave, demonstrating its ability to capture complex dynamics even for sharp solutions.

\subsection{Ablation Study}\label{subsec:abl-study}

\textbf{Number of Basis} The performance of SUPRA with different number of basis functions is compared. As presented in \cref{tab:number-of-basis}, it is worth noticing that SUPRA will not degenerate when the number of basis is small but also helps to filter out useless high-frequency modes.

\begin{table}[htbp]
    \centering
    \begin{tabular}{cc|ccc}
    \toprule
        \multirow{2}{*}{Test Case} & \multirow{2}{*}{Basis} &\multicolumn{3}{c}{\#Basis Functions} \\
        & & small & medium & large \\
    \midrule
        NS & Fourier & 6.86 & \textbf{6.32} & 6.50 \\
        Darcy & Fourier & \textbf{0.433} & 0.436 & 0.466 \\
        Darcy & Chebyshev & \textbf{0.432} & 0.449 & 0.454 \\
        Plasticity & Fourier & 0.062 & 0.056 & \textbf{0.044} \\
        Airfoil & Laplacian & 0.446 & \textbf{0.340} & 0.342 \\
    \bottomrule
    \end{tabular}
    \caption{Comparison between different numbers of basis functions: Relative $L^2$ error ($\times 10^{-2}$) is recorded. The numbers of basis functions range from $64$ to $256$: The small/medium/large corresponds to $64/100/144$ for Darcy and Plasticity, $64/144/256$ for Navier Stokes, and $64/128/192$ for Airfoil.}
    \label{tab:number-of-basis}
\end{table}

\textbf{Choice of Basis} As an example to construct the basis, on 1D domain $[0, 1]$, the Fourier basis with $n$ modes is:
\begin{equation}
    \{\cos(2 \pi i x), \sin(2\pi i x) \}_{i = 1}^{n}.
\end{equation}

We use a tensor product Fourier basis for the 2D domain $[0, 1]^2$. Assuming we have $m, n$ modes in the $x, y$ direction, the basis is constructed:
\begin{equation}
\begin{aligned}
    \{&\cos(2 \pi i x) \cos(2 \pi j y), \cos (2 \pi ix) \sin(2 \pi jy), \\
      &\sin(2 \pi i x) \cos(2 \pi j y), \sin (2 \pi ix) \sin(2 \pi jy)\}_{i,j=1}^{m, n}.
\end{aligned}
\end{equation}

For a general $d$-dimensional structured mesh, the mesh is first mapped to the $d$-dimensional unit square $[0, 1]^d$, and the tensor product basis on the unit square is constructed accordingly. A $d$-dimensional basis with $n$ modes per dimension results in $(2n)^d$ basis functions.

As mentioned in Section \cref{subsec:basis}, the choice of basis is problem dependent. The performance of the Fourier basis and Chebyshev basis is compared on Darcy, Navier-Stokes, and Plasticity problems in \cref{tab:comp-fourier-chebyshev}. 
\begin{table}[hbtp]
\centering
\begin{tabular}{c|ccc}
  \toprule
    \multirow{2}{*}{Basis} & \multicolumn{3}{c}{Test Case} \\
     & NS & Darcy & Plasticity \\
  \midrule
    Fourier & \textbf{6.32} & 0.466 & \textbf{0.044} \\
    Chebyshev & 6.60 & \textbf{0.454} & 0.049  \\
  \bottomrule
\end{tabular}
\caption{Comparison between different orthonormal basis on regular grids. The table records the relative $L^2$ error ($\times 10^{-2}$). The number of basis is set to $144$ in these experiments.}
\label{tab:comp-fourier-chebyshev}
\end{table}

The performance of the subspace spanned by Laplacian eigenfunctions and the Fourier basis is compared on the Airfoil problem, with experiment results listed in the \cref{tab:topo-naca-result}. We choose 128 Laplacian eigenfunctions and 6 modes for the Fourier basis (i.e. $(6\times2)^2=144$ basis functions in total). Since the eigenfunctions are computed directly on the irregular physical domain, the Laplacian eigensubspace outperforms a Fourier basis defined on a regular computational grid. 

\begin{table}[hbt]
    \centering
    \begin{tabular}{c|cc}
\toprule
    Dataset &  Laplacian & Fourier\\
\midrule
    Train & \textbf{0.74} &  0.93 \\
    Test &  \textbf{3.40} &  4.58 \\
\bottomrule
    \end{tabular}
    \caption{Comparison betteen Laplacian eigensubspace and Fourier basis. Relative $L^2$ Error ($\times 10^{-3}$) of Airfoil on train and test set is recorded with different subspace.}
    \label{tab:topo-naca-result}
\end{table}

\textbf{Normalization} The performance of LayerNorm and InstanceNorm is compared in Table \ref{tab:normalization}. It is clear that, the training process can fail without any normalization applied, while the performance of InstanceNorm is slightly better than LayerNorm.

\begin{table}[h]
    \centering
    \begin{tabular}{c|cccc}
    \toprule
        \multirow{2}{*}{Normalization} & \multicolumn{4}{c}{Test Case} \\
         & NS & Airfoil & Darcy & Plasticity \\
    \midrule
        Layer       & 6.92 & \textbf{0.340} & \textbf{0.433} & 0.046 \\
        Instance    & \textbf{6.50} & 0.350 & 0.463 & \textbf{0.044} \\
        None        & $\uparrow$ & $\uparrow$ & 0.479 & $\uparrow$ \\
    \bottomrule
    \end{tabular}
    \caption{Comparison between different normalizations. Relative $L^2$ error ($\times 10^{-2}$) is recorded in the table, and $\uparrow$ indicates the optimization process blows up.}
    \label{tab:normalization}
\end{table}

\subsection{Model Scalability}\label{subsec:model-scal}

\textbf{Model Size} The accuracy of the SUPRA neural operator is evaluated across different numbers of learnable parameters in \cref{tab:accuracy-layers}. Increasing the number of hidden features increases the number of functions that participate in attention while increasing the number of hidden layers $L$ enables the model to learn more complex features.

\begin{table}[h]
    \centering
    \begin{tabular}{c|ccc}
    \toprule
        \multirow{2}{*}{Model Size} & \multicolumn{3}{c}{Test Case} \\
         & Navier Stokes & Airfoil & Darcy\\
    \midrule
        $L=4$ & 6.89 & 0.510 & 0.512 \\
        $L=6$ & \textbf{6.25} & \textbf{0.340} & 0.471\\
        $L=8$ & 6.32 & 0.340 & \textbf{0.432}\\
    \hline
        Small & 11.1 & 0.484 & 0.488 \\
        Medium & 7.99 & \textbf{0.340} & 0.432 \\
        Large & \textbf{6.58} & 0.371 & \textbf{0.432} \\
    \bottomrule
    \end{tabular}
    \caption{Comparison between different number of hidden layers $L$ and hidden features. Relative $L^2$ error ($\times 10^{-2}$) is recorded in the table. For hidden features, Small/Medium/Large correspond to 64/128/196 for Darcy and Navier Stokes and 32/64/96 for Airfoil.}
    \label{tab:accuracy-layers}
\end{table}

\begin{table*}[h!]
    \centering
    \begin{tabular}{cc|cc|c|cc}
\toprule
        \multicolumn{2}{c|}{Hyper-parameters} &\multirow{2}{*}{Model} & \multirow{2}{*}{\#Params} & \multirow{2}{*}{GPU Mem (MB)} & \multicolumn{2}{c}{Time (sec.)} \\
    \#Hiddens & Mesh Size & &&& {Forward} & {Backward}  \\
\midrule
        \multirow{4}{*}{64} & \multirow{4}{*}{$64\times 64$} & Transolver & 712K & 483 & 0.017 & 0.040 \\
        & & LNO & 334K & 231 & 0.009 & 0.019 \\
        & & Ours ($N=128$) & 635K & 173 & 0.007 & 0.016 \\
        & & Ours ($N=256$) & 2.2M & 191 & 0.008 & 0.017 \\
    \hline
        \multirow{4}{*}{64} & \multirow{4}{*}{$128\times128$} & Transolver & 712K & 1822 & 0.063 & 0.148 \\
        & & LNO & 334K & 741 & 0.029 & 0.058 \\
        & & Ours ($N=128$) & 635K & 635 & 0.013 & 0.034 \\
        & & Ours ($N=256$) & 2.2M & 658 & 0.016 & 0.042 \\
    \hline
        \multirow{4}{*}{256}& \multirow{4}{*}{$64\times 64$} & Transolver & 11M & 1469 & 0.062 & 0.169  \\
        & & LNO & 5.0M & 683 & 0.017 & 0.040 \\
        & & Ours ($N=128$) & 2.1M & 662 & 0.015 & 0.040 \\
        & & Ours ($N=256$) & 3.7M & 708 & 0.019 & 0.047 \\
    \hline
        \multirow{4}{*}{256} & \multirow{4}{*}{$128\times 128$} & Transolver & 11M & 5647 & 0.236 & 0.677 \\
        & & LNO & 5.0M & 2141 & 0.051 & 0.121 \\
        & & Ours ($N=128$) & 2.1M & 2430 & 0.053 & 0.133 \\
        & & Ours ($N=256$) & 3.7M & 2479 & 0.071 & 0.177 \\
\bottomrule
    \end{tabular}
    \caption{Model efficiency. A single RTX-3060 is used for all experiments. All models and experiments use 8 hidden layers and a batch size of 4. $N$ represents the number of subspace basis functions in the SUPRA block.}
    \label{tab:profiler}
\end{table*}

\textbf{Computational Cost} 
The evaluation cost comparison of SUPRA and previous methods is listed in \cref{tab:profiler}. A 2D problem defined on a structured grid, with 1 function as input and 1 function as output, is considered in this experiment. By comparing \cref{tab:accuracy-layers} and \ref{tab:profiler}, it can be concluded that SUPRA maintains high prediction accuracy even with a small number of parameters and computational costs.

\section{Conclusion and Future Work}\label{sec:conclusion}

This paper introduces the SUPRA neural operator for solving PDEs on general domains. SUPRA defines attention between functions derived directly from the standard attention mechanism and parameterizes attention within a finite-dimensional subspace, achieving superior expressive power and computational efficiency. The SUPRA neural operator demonstrates stronger performance on irregular domains by leveraging the Laplacian eigensubspace defined directly on the physical domain. SUPRA achieves higher accuracy across various standard PDE benchmarks while maintaining low computational cost. In the future, manual subspace construction and choice can also be replaced by a machine learning model. Applying SUPRA to PDEs and 3D dynamics with a larger mesh is also challenging.

% \section*{Acknowledgements}
% We thank our coworkers Qing Fang for the technical insight on the Laplacian eigensubspace and Yan Xu for the computational resource. We also thank Ruojun Xu from Zhejiang University, Zhiyi Shi from Carnegie Mellon University and Rui Xu from Southeast University for their useful feedback on the manuscript.

\section*{Impact Statement}
This work aims to improve neural networks as surrogate models for PDE solvers by introducing the attention mechanism directly in function space. This approach offers superior efficiency and expressive power compared to current techniques. We are confident that our method can enhance performance in practical applications, including weather prediction and topology optimization. In developing our approach, we have prioritized ethical considerations and ensured that our work carries no foreseeable ethical concerns.


\bibliography{example_paper}

\begin{thebibliography}{46}
\providecommand{\natexlab}[1]{#1}
\providecommand{\url}[1]{\texttt{#1}}
\expandafter\ifx\csname urlstyle\endcsname\relax
  \providecommand{\doi}[1]{doi: #1}\else
  \providecommand{\doi}{doi: \begingroup \urlstyle{rm}\Url}\fi

\bibitem[Ba et~al.(2016)Ba, Kiros, and Hinton]{ba2016layern}
Ba, J., Kiros, J.~R., and Hinton, G.~E.
\newblock Layer normalization.
\newblock \emph{ArXiv}, abs/1607.06450, 2016.

\bibitem[Benner et~al.(2015)Benner, Gugercin, and
  Willcox]{siam2015projmodelreduction}
Benner, P., Gugercin, S., and Willcox, K.
\newblock A survey of projection-based model reduction methods for parametric
  dynamical systems.
\newblock \emph{SIAM Review}, 57\penalty0 (4):\penalty0 483--531, 2015.
\newblock \doi{10.1137/130932715}.

\bibitem[Benner et~al.(2017)Benner, Ohlberger, Cohen, and
  Willcox]{mor2017survey}
Benner, P., Ohlberger, M., Cohen, A., and Willcox, K.
\newblock \emph{Model Reduction and Approximation}.
\newblock Society for Industrial and Applied Mathematics, Philadelphia, PA,
  2017.
\newblock \doi{10.1137/1.9781611974829}.

\bibitem[Bernardi \& Maday(1997)Bernardi and Maday]{bernardi1997spectral}
Bernardi, C. and Maday, Y.
\newblock Spectral methods.
\newblock In \emph{Techniques of Scientific Computing (Part 2)}, volume~5 of
  \emph{Handbook of Numerical Analysis}, pp.\  209--485. Elsevier, 1997.

\bibitem[Bonev et~al.(2023)Bonev, Kurth, Hundt, Pathak, Baust, Kashinath, and
  Anandkumar]{bonev2023sphericalfn}
Bonev, B., Kurth, T., Hundt, C., Pathak, J., Baust, M., Kashinath, K., and
  Anandkumar, A.
\newblock Spherical fourier neural operators: Learning stable dynamics on the
  sphere.
\newblock In \emph{International Conference on Machine Learning}, 2023.

\bibitem[Brenner \& Scott(2008)Brenner and Scott]{brenner2008fem}
Brenner, S. and Scott, R.
\newblock \emph{The Mathematical Theory of Finite Element Method}, volume~15.
\newblock 01 2008.
\newblock ISBN 978-1-4757-4340-1.
\newblock \doi{10.1007/978-1-4757-4338-8}.

\bibitem[Brunton et~al.(2019)Brunton, Noack, and
  Koumoutsakos]{brunton2019machinelf}
Brunton, S.~L., Noack, B.~R., and Koumoutsakos, P.
\newblock Machine learning for fluid mechanics.
\newblock \emph{ArXiv}, abs/1905.11075, 2019.

\bibitem[Cao(2021)]{cao2021chooseat}
Cao, S.
\newblock Choose a transformer: Fourier or galerkin.
\newblock In \emph{Neural Information Processing Systems}, 2021.

\bibitem[Choi et~al.(2020)Choi, Boncoraglio, Anderson, Amsallem, and
  Farhat]{choi2020gradientbased}
Choi, Y., Boncoraglio, G., Anderson, S., Amsallem, D., and Farhat, C.
\newblock Gradient-based constrained optimization using a database of linear
  reduced-order models.
\newblock \emph{Journal of Computational Physics}, 423:\penalty0 109787, 2020.
\newblock ISSN 0021-9991.
\newblock \doi{https://doi.org/10.1016/j.jcp.2020.109787}.

\bibitem[De~Witt et~al.(2012)De~Witt, Lessig, and
  Fiume]{witt2012fluidlaplacian}
De~Witt, T., Lessig, C., and Fiume, E.
\newblock Fluid simulation using laplacian eigenfunctions.
\newblock \emph{ACM Trans. Graph.}, 31\penalty0 (1), February 2012.
\newblock ISSN 0730-0301.
\newblock \doi{10.1145/2077341.2077351}.

\bibitem[Evans(2010)]{evans2010partial}
Evans, L.
\newblock \emph{Partial Differential Equations}.
\newblock Graduate studies in mathematics. American Mathematical Society, 2010.
\newblock ISBN 9780821849743.

\bibitem[Gilbarg \& Trudinger(2013)Gilbarg and Trudinger]{gilbarg2013elliptic}
Gilbarg, D. and Trudinger, N.
\newblock \emph{Elliptic Partial Differential Equations of Second Order}.
\newblock Grundlehren der mathematischen Wissenschaften. Springer Berlin
  Heidelberg, 2013.
\newblock ISBN 9783642963797.

\bibitem[Hao et~al.(2023)Hao, Wang, Su, Ying, Dong, Liu, Cheng, Song, and
  Zhu]{hao2023gnot}
Hao, Z., Wang, Z., Su, H., Ying, C., Dong, Y., Liu, S., Cheng, Z., Song, J.,
  and Zhu, J.
\newblock Gnot: A general neural operator transformer for operator learning.
\newblock In \emph{International Conference on Machine Learning, ICML 2023,
  23-29 July 2023, Honolulu, Hawaii, USA}, volume 202 of \emph{Proceedings of
  Machine Learning Research}, pp.\  12556--12569. PMLR, 2023.

\bibitem[Hesthaven(2018)]{hesthaven2018numericalmethod}
Hesthaven, J.~S.
\newblock \emph{Numerical Methods for Conservation Laws}.
\newblock Society for Industrial and Applied Mathematics, Philadelphia, PA,
  2018.
\newblock \doi{10.1137/1.9781611975109}.

\bibitem[Horie \& Mitsume(2024)Horie and Mitsume]{horie2024conservation}
Horie, M. and Mitsume, N.
\newblock Graph neural {PDE} solvers with conservation and
  similarity-equivariance.
\newblock In Salakhutdinov, R., Kolter, Z., Heller, K., Weller, A., Oliver, N.,
  Scarlett, J., and Berkenkamp, F. (eds.), \emph{Proceedings of the 41st
  International Conference on Machine Learning}, volume 235 of
  \emph{Proceedings of Machine Learning Research}, pp.\  18785--18814. PMLR,
  21--27 Jul 2024.

\bibitem[Karniadakis et~al.(2021)Karniadakis, Kevrekidis, Lu, Perdikaris, Wang,
  and Yang]{karniadakis2021physicsinformedml}
Karniadakis, G.~E., Kevrekidis, I.~G., Lu, L., Perdikaris, P., Wang, S., and
  Yang, L.
\newblock Physics-informed machine learning.
\newblock \emph{Nature Reviews Physics}, 3:\penalty0 422 -- 440, 2021.

\bibitem[Kossaifi et~al.(2023)Kossaifi, Kovachki, Azizzadenesheli, and
  Anandkumar]{kossaifi2023multigridtf}
Kossaifi, J., Kovachki, N.~B., Azizzadenesheli, K., and Anandkumar, A.
\newblock Multi-grid tensorized fourier neural operator for high-resolution
  pdes.
\newblock \emph{ArXiv}, abs/2310.00120, 2023.

\bibitem[Kovachki et~al.(2023)Kovachki, Li, Liu, Azizzadenesheli, Bhattacharya,
  Stuart, and Anandkumar]{kovachki2023neuraloperator}
Kovachki, N., Li, Z., Liu, B., Azizzadenesheli, K., Bhattacharya, K., Stuart,
  A., and Anandkumar, A.
\newblock Neural operator: Learning maps between function spaces with
  applications to pdes.
\newblock \emph{Journal of Machine Learning Research}, 24\penalty0
  (89):\penalty0 1--97, 2023.

\bibitem[L\'{e}vy \& Zhang(2010)L\'{e}vy and
  Zhang]{levy2010spectralmeshprocessing}
L\'{e}vy, B. and Zhang, H.~R.
\newblock Spectral mesh processing.
\newblock In \emph{ACM SIGGRAPH 2010 Courses}, SIGGRAPH '10, New York, NY, USA,
  2010. Association for Computing Machinery.
\newblock ISBN 9781450303958.
\newblock \doi{10.1145/1837101.1837109}.

\bibitem[Li et~al.(2022{\natexlab{a}})Li, Meidani, and Farimani]{li2022oformer}
Li, Z., Meidani, K., and Farimani, A.~B.
\newblock Transformer for partial differential equations' operator learning.
\newblock \emph{Trans. Mach. Learn. Res.}, 2023, 2022{\natexlab{a}}.

\bibitem[Li et~al.(2023{\natexlab{a}})Li, Shu, and
  Barati~Farimani]{li2023factformer}
Li, Z., Shu, D., and Barati~Farimani, A.
\newblock Scalable transformer for pde surrogate modeling.
\newblock In Oh, A., Naumann, T., Globerson, A., Saenko, K., Hardt, M., and
  Levine, S. (eds.), \emph{Advances in Neural Information Processing Systems},
  volume~36, pp.\  28010--28039. Curran Associates, Inc., 2023{\natexlab{a}}.

\bibitem[Li et~al.(2020)Li, Kovachki, Azizzadenesheli, Liu, Bhattacharya,
  Stuart, and Anandkumar]{li2020fourierno}
Li, Z.-Y., Kovachki, N.~B., Azizzadenesheli, K., Liu, B., Bhattacharya, K.,
  Stuart, A.~M., and Anandkumar, A.
\newblock Fourier neural operator for parametric partial differential
  equations.
\newblock \emph{ArXiv}, abs/2010.08895, 2020.

\bibitem[Li et~al.(2022{\natexlab{b}})Li, Huang, Liu, and
  Anandkumar]{li2022geofno}
Li, Z.-Y., Huang, D.~Z., Liu, B., and Anandkumar, A.
\newblock Fourier neural operator with learned deformations for pdes on general
  geometries.
\newblock \emph{J. Mach. Learn. Res.}, 24:\penalty0 388:1--388:26,
  2022{\natexlab{b}}.

\bibitem[Li et~al.(2023{\natexlab{b}})Li, Kovachki, Choy, Li, Kossaifi, Otta,
  Nabian, Stadler, Hundt, Azizzadenesheli, and
  Anandkumar]{li2023geometryinformedno}
Li, Z.-Y., Kovachki, N.~B., Choy, C., Li, B., Kossaifi, J., Otta, S.~P.,
  Nabian, M.~A., Stadler, M., Hundt, C., Azizzadenesheli, K., and Anandkumar,
  A.
\newblock Geometry-informed neural operator for large-scale 3d pdes.
\newblock \emph{ArXiv}, abs/2309.00583, 2023{\natexlab{b}}.

\bibitem[Lingsch et~al.(2024)Lingsch, Michelis, De~Bezenac, M.~Perera,
  Katzschmann, and Mishra]{lingsch2024beyoundregulargrids}
Lingsch, L.~E., Michelis, M.~Y., De~Bezenac, E., M.~Perera, S., Katzschmann,
  R.~K., and Mishra, S.
\newblock Beyond regular grids: {F}ourier-based neural operators on arbitrary
  domains.
\newblock In Salakhutdinov, R., Kolter, Z., Heller, K., Weller, A., Oliver, N.,
  Scarlett, J., and Berkenkamp, F. (eds.), \emph{Proceedings of the 41st
  International Conference on Machine Learning}, volume 235 of
  \emph{Proceedings of Machine Learning Research}, pp.\  30610--30629. PMLR,
  21--27 Jul 2024.

\bibitem[Loshchilov \& Hutter(2017)Loshchilov and
  Hutter]{Loshchilov2017DecoupledWD}
Loshchilov, I. and Hutter, F.
\newblock Decoupled weight decay regularization.
\newblock In \emph{International Conference on Learning Representations}, 2017.

\bibitem[Lu et~al.(2019)Lu, Jin, Pang, Zhang, and
  Karniadakis]{lu2019learningno}
Lu, L., Jin, P., Pang, G., Zhang, Z., and Karniadakis, G.~E.
\newblock Learning nonlinear operators via deeponet based on the universal
  approximation theorem of operators.
\newblock \emph{Nature Machine Intelligence}, 3:\penalty0 218 -- 229, 2019.

\bibitem[Rahman et~al.(2024)Rahman, George, Elleithy, Leibovici, Li, Bonev,
  White, Berner, Yeh, Kossaifi, Azizzadenesheli, and
  Anandkumar]{rahman2024pretraining}
Rahman, M.~A., George, R.~J., Elleithy, M., Leibovici, D., Li, Z., Bonev, B.,
  White, C., Berner, J., Yeh, R.~A., Kossaifi, J., Azizzadenesheli, K., and
  Anandkumar, A.
\newblock Pretraining codomain attention neural operators for solving
  multiphysics pdes.
\newblock \emph{Advances in Neural Information Processing Systems}, 37, 2024.

\bibitem[Raissi et~al.(2017)Raissi, Perdikaris, and
  Karniadakis]{raissi2017physicsi}
Raissi, M., Perdikaris, P., and Karniadakis, G.~E.
\newblock Physics informed deep learning (part i): Data-driven solutions of
  nonlinear partial differential equations.
\newblock \emph{arXiv preprint arXiv:1711.10561}, 2017.

\bibitem[Raissi et~al.(2019)Raissi, Perdikaris, and
  Karniadakis]{raissi2019physics}
Raissi, M., Perdikaris, P., and Karniadakis, G.~E.
\newblock Physics-informed neural networks: A deep learning framework for
  solving forward and inverse problems involving nonlinear partial differential
  equations.
\newblock \emph{Journal of Computational Physics}, 378:\penalty0 686--707,
  2019.

\bibitem[Rathore et~al.(2024)Rathore, Lei, Frangella, Lu, and
  Udell]{rathore2024pinnchallenges}
Rathore, P., Lei, W., Frangella, Z., Lu, L., and Udell, M.
\newblock Challenges in training {PINN}s: A loss landscape perspective.
\newblock In Salakhutdinov, R., Kolter, Z., Heller, K., Weller, A., Oliver, N.,
  Scarlett, J., and Berkenkamp, F. (eds.), \emph{Proceedings of the 41st
  International Conference on Machine Learning}, volume 235 of
  \emph{Proceedings of Machine Learning Research}, pp.\  42159--42191. PMLR,
  21--27 Jul 2024.

\bibitem[Sheng et~al.(2021)Sheng, Cao, and Shen]{sheng2021spectral}
Sheng, C., Cao, D., and Shen, J.
\newblock Efficient spectral methods for pdes with spectral fractional
  laplacian.
\newblock \emph{Journal of Scientific Computing}, 88, 07 2021.
\newblock \doi{10.1007/s10915-021-01491-2}.

\bibitem[Sifakis \& Barbic(2012)Sifakis and Barbic]{sifakis2012femsimulation3d}
Sifakis, E. and Barbic, J.
\newblock Fem simulation of 3d deformable solids: a practitioner's guide to
  theory, discretization and model reduction.
\newblock In \emph{ACM SIGGRAPH 2012 Courses}, SIGGRAPH '12, New York, NY, USA,
  2012. Association for Computing Machinery.
\newblock ISBN 9781450316781.
\newblock \doi{10.1145/2343483.2343501}.

\bibitem[Sirovich(1987)]{sirovich1987turbulenceat}
Sirovich, L.
\newblock Turbulence and the dynamics of coherent structures. i. coherent
  structures.
\newblock \emph{Quarterly of Applied Mathematics}, 45:\penalty0 561--571, 1987.

\bibitem[Smith et~al.(1996)Smith, Bj\o{}rstad, and Gropp]{domaindecomposition}
Smith, B.~F., Bj\o{}rstad, P.~E., and Gropp, W.~D.
\newblock \emph{Domain decomposition: parallel multilevel methods for elliptic
  partial differential equations}.
\newblock Cambridge University Press, USA, 1996.
\newblock ISBN 052149589X.

\bibitem[Smith \& Topin(2018)Smith and Topin]{Smith2018SuperconvergenceVF}
Smith, L.~N. and Topin, N.
\newblock Super-convergence: very fast training of neural networks using large
  learning rates.
\newblock In \emph{Defense + Commercial Sensing}, 2018.

\bibitem[Stein \& Shakarchi(2009)Stein and Shakarchi]{stein2009real}
Stein, E. and Shakarchi, R.
\newblock \emph{Real Analysis: Measure Theory, Integration, and Hilbert
  Spaces}.
\newblock Princeton University Press, 2009.
\newblock ISBN 9781400835560.

\bibitem[Trefethen(2013)]{chebyshev_basis}
Trefethen, L.~N.
\newblock \emph{Approximation theory and approximation practice}.
\newblock Society for Industrial and Applied Mathematics (SIAM), Philadelphia,
  PA, 2013.
\newblock ISBN 978-1-611972-39-9.

\bibitem[Ulyanov et~al.(2016)Ulyanov, Vedaldi, and
  Lempitsky]{ul2016instancenorm}
Ulyanov, D., Vedaldi, A., and Lempitsky, V.~S.
\newblock Instance normalization: The missing ingredient for fast stylization.
\newblock \emph{CoRR}, abs/1607.08022, 2016.

\bibitem[Vaswani et~al.(2017)Vaswani, Shazeer, Parmar, Uszkoreit, Jones, Gomez,
  Kaiser, and Polosukhin]{vaswani2017attentionia}
Vaswani, A., Shazeer, N.~M., Parmar, N., Uszkoreit, J., Jones, L., Gomez,
  A.~N., Kaiser, L., and Polosukhin, I.
\newblock Attention is all you need.
\newblock In \emph{Neural Information Processing Systems}, 2017.

\bibitem[Wang et~al.(2024)Wang, Jiaxin, Dwivedi, Hara, and Wu]{wang2024beno}
Wang, H., Jiaxin, L., Dwivedi, A., Hara, K., and Wu, T.
\newblock {BENO}: Boundary-embedded neural operators for elliptic {PDE}s.
\newblock In \emph{The Twelfth International Conference on Learning
  Representations}, 2024.

\bibitem[Wang \& Wang(2024)Wang and Wang]{wang2024lno}
Wang, T. and Wang, C.
\newblock Latent neural operator for solving forward and inverse pde problems.
\newblock In \emph{Advances in Neural Information Processing Systems}, 2024.

\bibitem[Wen et~al.(2022)Wen, Li, Azizzadenesheli, Anandkumar, and
  Benson]{wen2022ufno}
Wen, G., Li, Z., Azizzadenesheli, K., Anandkumar, A., and Benson, S.~M.
\newblock U-fno—an enhanced fourier neural operator-based deep-learning model
  for multiphase flow.
\newblock \emph{Advances in Water Resources}, 163:\penalty0 104180, 2022.
\newblock ISSN 0309-1708.
\newblock \doi{https://doi.org/10.1016/j.advwatres.2022.104180}.

\bibitem[Wright et~al.(2015)Wright, Javed, Montanelli, and
  Trefethen]{fourier_basis}
Wright, G.~B., Javed, M., Montanelli, H., and Trefethen, L.~N.
\newblock Extension of {C}hebfun to periodic functions.
\newblock \emph{SIAM J. Sci. Comput.}, 37\penalty0 (5):\penalty0 C554--C573,
  2015.
\newblock ISSN 1064-8275,1095-7197.
\newblock \doi{10.1137/141001007}.
\newblock URL \url{https://doi.org/10.1137/141001007}.

\bibitem[Wu et~al.(2024)Wu, Luo, Wang, Wang, and Long]{wu2024transolveraf}
Wu, H., Luo, H., Wang, H., Wang, J., and Long, M.
\newblock Transolver: A fast transformer solver for {PDE}s on general
  geometries.
\newblock In Salakhutdinov, R., Kolter, Z., Heller, K., Weller, A., Oliver, N.,
  Scarlett, J., and Berkenkamp, F. (eds.), \emph{Proceedings of the 41st
  International Conference on Machine Learning}, volume 235 of
  \emph{Proceedings of Machine Learning Research}, pp.\  53681--53705. PMLR,
  21--27 Jul 2024.

\bibitem[Xiao et~al.(2023)Xiao, Hao, Lin, Deng, and Su]{xiao2023improvedol}
Xiao, Z., Hao, Z., Lin, B., Deng, Z., and Su, H.
\newblock Improved operator learning by orthogonal attention.
\newblock \emph{ArXiv}, abs/2310.12487, 2023.

\end{thebibliography}
\bibliographystyle{icml2025}

%%%%%%%%%%%%%%%%%%%%%%%%%%%%%%%%%%%%%%%%%%%%%%%%%%%%%%%%%%%%%%%%%%%%%%%%%%%%%%%
%%%%%%%%%%%%%%%%%%%%%%%%%%%%%%%%%%%%%%%%%%%%%%%%%%%%%%%%%%%%%%%%%%%%%%%%%%%%%%%
% APPENDIX
%%%%%%%%%%%%%%%%%%%%%%%%%%%%%%%%%%%%%%%%%%%%%%%%%%%%%%%%%%%%%%%%%%%%%%%%%%%%%%%
%%%%%%%%%%%%%%%%%%%%%%%%%%%%%%%%%%%%%%%%%%%%%%%%%%%%%%%%%%%%%%%%%%%%%%%%%%%%%%%
\newpage
\appendix
\onecolumn
% \section{Bilinear Forms}\label{sec:bilinear}
% \subsection{Numerical PDE solvers}\label{subsec:numerical-pde-bilinear}

\section{Approximation property of SUPRA}\label{subsec:approx-bilinear-bounded}
In this section, we examine SUPRA will converge under the assumption of boundness of the linear operator $b(\cdot)$ and bilinear form $a(\cdot, \cdot)$. In a Banach space, we call a linear operator bounded if there is a constant $C$ such that for all $u \in \mathbf{V}$,
\begin{equation}
    \|b(u)\| \le C\cdot \|u\|.
\end{equation}
A bilinear form is bounded if there is a constant $C$ such that for all $u, v\in \mathbf{V}$,
\begin{equation}
    a(u, v) \le C \cdot \| u\| \cdot \|v\|.
\end{equation}

The boundness property of linear operators and bilinear forms are identical to continuity in Banach spaces: Suppose $u_i$ and $v_i$ converge to $u$ and $v$ in space $\mathbf{V}$, for linear operator $b(\cdot)$,
\begin{equation}
\begin{aligned}
     &\lim_{i\to \infty} \|u_i - u\| = 0
\implies & 0 \le \lim_{i\to\infty} \|b(u_i) - b(u)\| 
= \lim_{i\to \infty} \| b(u_i - u) \| \le C \lim_{i\to \infty} \|u_i - u\| = 0.
\end{aligned}
\end{equation}
Therefore, $\lim_{i\to\infty} b(u_i) = b(u)$. Similarly, we have $\lim_{i, j\to\infty}a(u_i, v_j) = a(u, v)$ as well. Suppose we use $N$ basis functions $\{e_i\}_{i=1}^N$ to approximate $u, v$, as $N$ goes to infinity:
\begin{equation}
    \lim_{N\to \infty} \sum_{k=1}^N \hat{u}^k e_k = u, \quad \lim_{N\to \infty} \sum_{k=1}^N \hat{v}^k e_k = v.
\end{equation}
Substituting the summation into \cref{eq:attn-parameterized}, we have:
\begin{equation}
\begin{aligned}
    \lim_{N \to \infty} a\left( \sum_{k=1}^N \hat{u}^k e_k, \sum_{k=1}^N \hat{v}^k e_k \right)&= a(u, v), \\
    \lim_{N \to \infty} b\left( \sum_{k=1}^N \hat{u}^k e_k \right) &= b(u).
\end{aligned}
\end{equation}
Therefore, as $N$ increases, SUPRA's outputs will converge to the attention of the vector space.

\textbf{Subspace Basis} Here, we list some common basis. These basis share a property that, as the number of basis functions $N$ increases, the approximations $u_N = \sum_{i=1}^N \hat{u}^k e_k$ will converge to $u$. These properties guarantee the existence of $u_i, v_i$ in the proof above.

\begin{enumerate}
    \item Fourier/Trigonometric representations have uniform properties across the interval of approximation, which is also the Laplacian eigenfunctions on regular grids such as $[0, 1]^d$,
    \item Chebyshev polynomials are nonuniform, with greater resolution at the ends of $[-1, 1]$ than at the middle,
    % \item Chebyshev polynomials are nonuniform, with much greater resolution power near the ends of  $[-1, 1]$ than near the middle.
    \item Spherical harmonics are the natural extension of the Fourier series to functions defined on the surface of a sphere, which are also the Laplacian eigenfunctions on that surface,
\end{enumerate}

\section{Eigensubspace of Laplace Operator}\label{sec:topo-change}

\textbf{Benefits of the Eigensubspace} Laplace operator (Laplace-Beltrami operator, or Laplacian) is defined as the divergence of gradient $\Delta f =\mathrm{div} (\nabla f)$. The spectrum of the Laplacian consists of all eigenvalues $\lambda$ and eigenfunctions $f$ with:
\begin{equation}
    -\Delta f = \lambda f.
\end{equation}
This is also known as \textbf{Helmholtz Equation}. If the domain $\omega$ is bounded in $\mathbb{R}^n$, then eigenfunctions of the Laplacian are an orthonormal basis for Hilbert space $L^2(\Omega)$ \cite{gilbarg2013elliptic}. \textbf{Every eigenfunction is infinitely differentiable, guaranteeing continuity across any irregular domain.} Besides, the theorem below describes the smallest eigenfunctions as the most smooth functions on $\Omega$ \cite{evans2010partial}:
\begin{theorem}
    The infimum is achieved if and only if $\varphi$ is an eigenfunction of eigenvalue $\lambda_k$:
    \begin{equation}
        \lambda_k = \inf \left\{ \frac{\|\nabla \varphi\|^2_{L^2}}{\|\varphi\|^2_{L^2}}: s.t. \varphi \in H_1(\Omega) \cap E_k \right \}
    \end{equation}
    where $E_k:=\{\varphi_1, \varphi_2, \cdots, \varphi_{k-1}\}^\bot$.
\end{theorem}

Theory and application of the discrete version of Laplacian eigensubspace are also well-established in spectral mesh processing \cite{levy2010spectralmeshprocessing}. This eigensubspace is used to perform effective and information-preserving model reduction while revealing global and intrinsic structures in geometric data, enabling efficient PDE solving \cite{sheng2021spectral} and simulation \cite{witt2012fluidlaplacian}.

% \textbf{Drawbacks} The eigensubspace of the laplace operator does have drawbacks. Generally speaking, we assume the topology of input does not change, and the points' location does not change much to reuse the eigensubspace (see our approach to construct the eigensubspace for NACA Airfoil in Figure \ref{fig:naca-mean-mesh-vs}). For a point cloud input (e.g. the Elasticity problem \cite{li2022geofno}), the topology of the input mesh may change significantly, so constructing a proper subspace will be more difficult.\label{par:drawback-eigensubspace}

\section{Experiments Setups}\label{sec:experiments-full}

\subsection{PDEs Descriptions and Data Preprocessing}
Consider the boundary-value problem given by:
\begin{equation}
\begin{aligned}
    L u = f, \quad &\text{in } \Omega, \\
      u = g, \quad &\text{on } \partial \Omega.
\end{aligned}
\end{equation}
Our experiments cover most variables in the form: (1) coefficients in $L$, describing the property of the material at different positions, (2) external force $f$, (3) domain shape $\Omega$, and (4) boundary shape $\partial \Omega$. Our experiments also consider time-dependent problems, which occur when previous states determine future states. We list our experiments in Table \ref{tab:benchmark-cases}.
\begin{table}[h]
\centering
\begin{tabular}{c|c|c|c|cc|c}
  \toprule
    \multirow{2}{*}{Test Case} & \multirow{2}{*}{Geometry} & \multirow{2}{*}{\#Dim} & \multirow{2}{*}{Mesh Size} & \multicolumn{2}{c}{Dataset Split} & \multirow{2}{*}{Input Type} \\
    & & & & Train & Test & \\
  \midrule
    Darcy \cite{li2020fourierno} &  Regular Grid & 2 & $85\times85$ & 1000 & 200 & Coefficients \\
    Navier Stokes \cite{li2020fourierno} & Regular Grid & 2+1 & $64\times64$ & 1000 & 200 & Previous States \\
    Plasticity \cite{li2022geofno} & Structured Mesh & 2+1 & $101\times31$ & 900 & 80 & External Force \\
    Airfoil \cite{li2022geofno} & Structured Mesh\textsuperscript{*} & 2 & $221\times 51$ & 1000 & 200 & Boundary Shape \\
    Pipe \cite{li2022geofno} & Structured Mesh & 2 & $129\times129$ & 1000 & 200 & Domain Shape \\
  \bottomrule
\end{tabular}
\caption{Datasets and their split used in our experiments. Although the Airfoil case is structured, the mesh is cut to map from an annulus-like domain to a regular grid. To construct the Laplacian eigensubspace, we regard it as an irregular mesh.}
\label{tab:benchmark-cases}
\end{table}

\textbf{Darcy} The equation in Darcy describes the flow of fluid through a porous medium:
\begin{equation}
\begin{aligned}
    - \nabla \cdot (a(x) \nabla u(x)) &= f(x), &\quad x&\in(0,1)^2,\\
    u(x)&=0,&\quad x&\in \partial (0,1)^2,
\end{aligned}
\end{equation}
where $a(x)$ is the diffusion coefficient of a porous medium, which is the input of neural operators and is stored by a regular grid. The output is the solution $u$ on the same grid. In this dataset, the original resolution is $421\times421$. We perform a $5\times$ downsample on the original one as previous works do. We further enhance the dataset by applying randomly flip horizontally or vertically to both input and output since the equation guarantees the operation. 

\textbf{Navier Stokes} This equation models incompressible and viscous flow with periodic boundary conditions:
\begin{equation}
\begin{aligned}
\partial_t w(x, t) + u(x, t) \cdot \nabla w(x, t) &= \nu \nabla w(x,t) + f(x), & \quad x&\in(0,1)^2, t \in (0, T],\\
\nabla \cdot u(x,t) & = 0, & \quad x&\in(0,1)^2, t \in (0, T],\\
w(x,0) &= w_0(x), & \quad x &\in (0, 1)^2,
\end{aligned}
\end{equation}
where $ u$ is the velocity and $ w=\nabla \times u$ is the vorticity of the flow. We choose $\nu = 10^{-5}$ in this case. Each sample in the dataset has the vorticity field in 20 time steps. The neural operator takes 10 steps as input and predicts the next 1 step. It advances in time by dropping the first step in history and appending the prediction of the neural operator.

\textbf{Plasticity} In this task, neural operators need to predict future deformation of a given plasticity material under the different impacts from above. Since the initial state of the plasticity does not change, the neural operators take time $t$, boundary condition on the top as input, and directly predict the deformation field in the domain.

\textbf{Pipe} In this task, neural operators need to predict the $x$-component of the velocity field of the fluid in pipes with different shapes. The input is the structured mesh's physical position under the pipe's deformation, and the output is the $x$-component of the velocity field on the same structured mesh.

\begin{figure}[htbp]
    \centering
    \subfigure[Plasticity]{\includegraphics[width=0.39\linewidth]{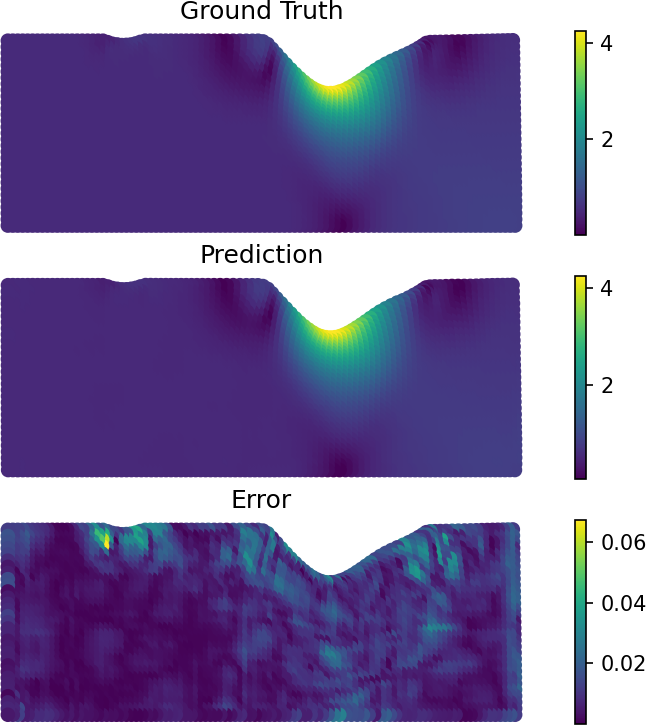}}
    \subfigure[Pipe]{\includegraphics[width=0.4\linewidth]{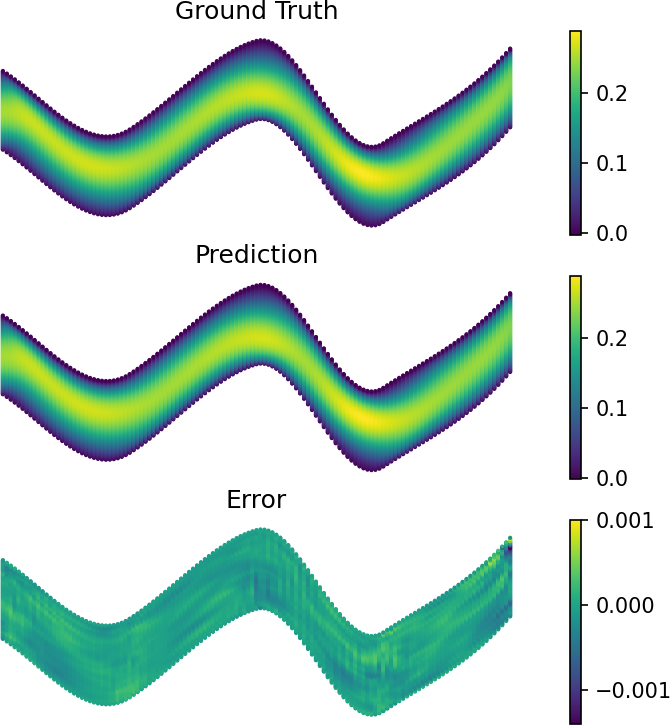}}
    \caption{Comparison between ground truth in Plasticity and Pipe test set.}
    \label{fig:plast-and-pipe}
\end{figure}

\textbf{Airfoil} Euler equation models the transonic flow over an airfoil. The input is mesh point locations, and the output is the Mach number on a structured mesh. We extract the average shape as the mean mesh to precompute the Laplacian eigensubspace. We compute the eigensubspace for only one mesh (i.e. the mean mesh), not for each sample.
\begin{figure}[htbp]
    \centering
    \subfigure[Comparison between mean shape and one example.]{\includegraphics[width=0.3\linewidth]{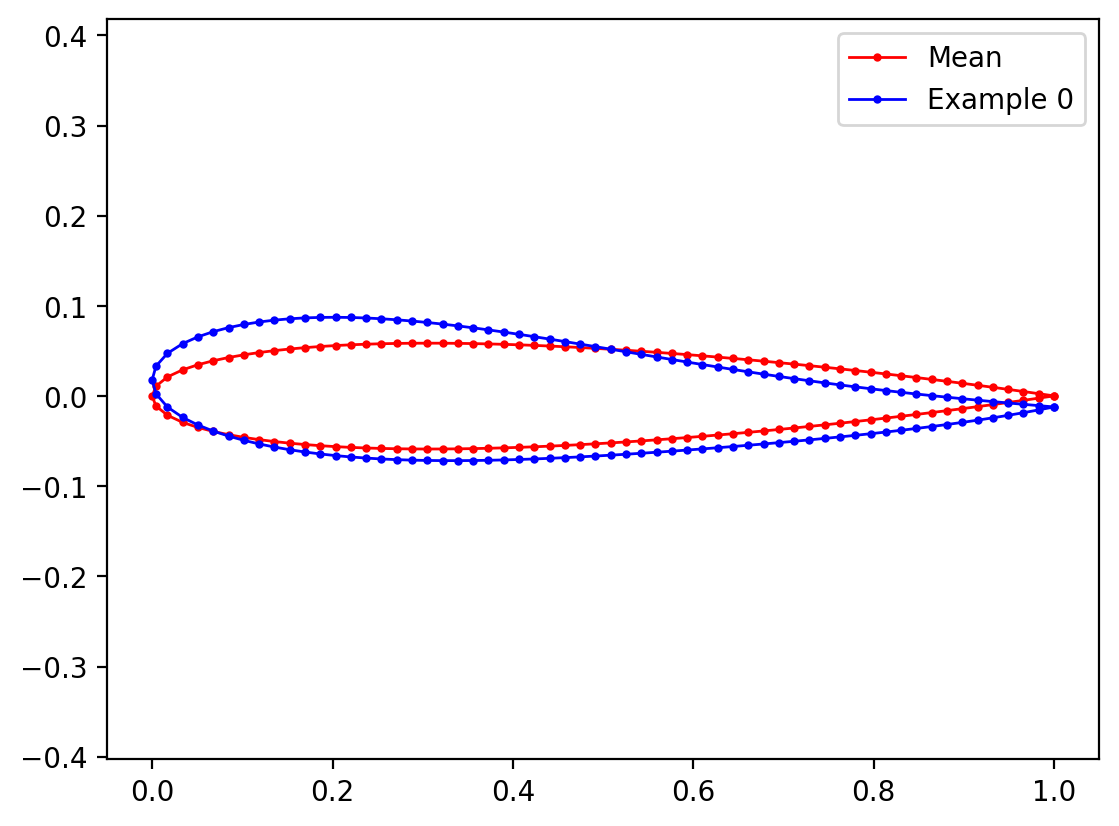}}
    \subfigure[Finite Element discretization on mean mesh.]{
    \includegraphics[width=0.3\linewidth]{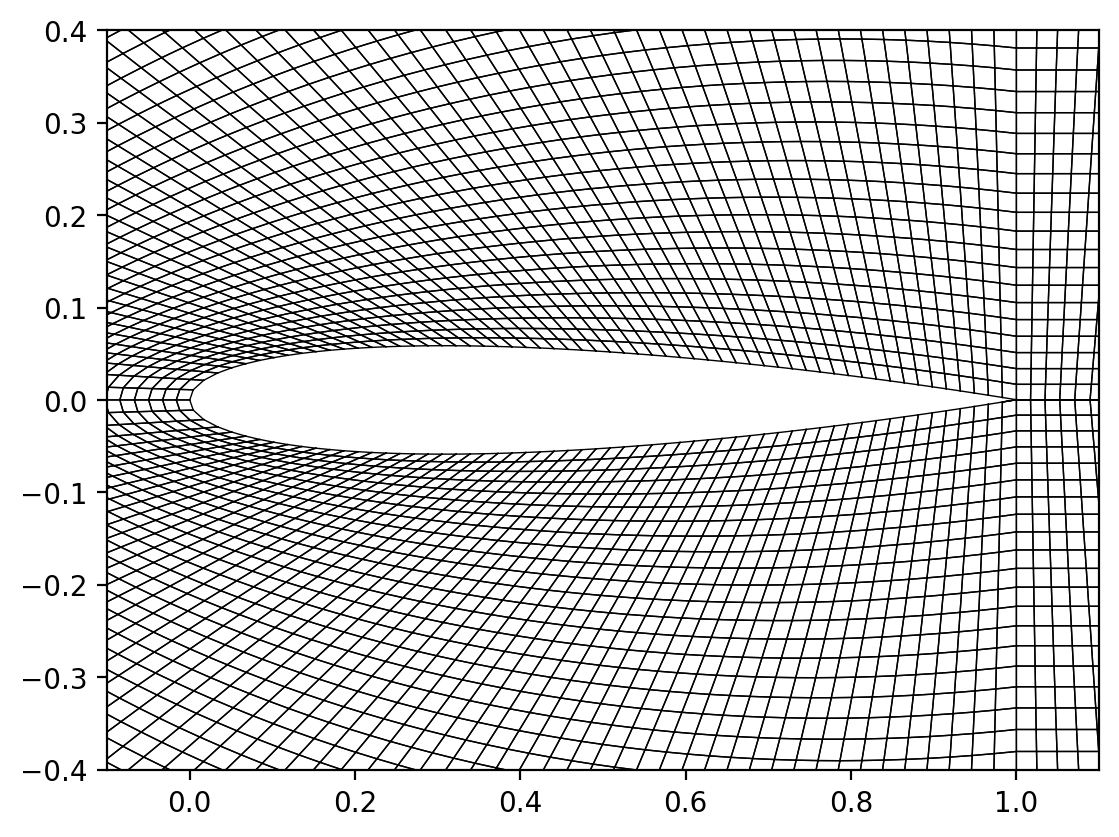}}
    \caption{Laplacian eigensubspace construction in Airfoil.}
    \label{fig:naca-mean-mesh-vs}
\end{figure}

We use input/output normalizers in each case, which is a common strategy to guarantee the stability of training progress. We use AdamW optimizer\cite{Loshchilov2017DecoupledWD} and a OneCycle scheduler\cite{Smith2018SuperconvergenceVF} for all the test cases.

\subsection{Configurations for optimal records}\label{subsec:optimal-conf}

We list the configuration of optimal records in \cref{tab:optimal-configuration}. In the table, $H_1$ loss corresponds to the Sobolev seminorm in the function space, which measures the regularity of the function. In our experiments on Darcy, $|u|_{H_1} = \|\nabla u\|_{L_2}$. Besides, all the losses in the table are relative. Different time steps in a data sample are considered multiple samples in training.

\begin{table}[htbp]
    \centering
    % \begin{tabular}{c|cc|ccccc}
    % \toprule
    %     Test Case & Batch Size & Loss Fn & Basis & \#Basis & Normalization & \#Layers & \#Hiddens \\
    % \midrule
    %     Darcy & 4 & $L_2 + 0.1 H_1$ & Chebyshev & 100 & Layer & 8 & 128 \\
    %     Navier Stokes & 20 & $L_2$ & Fourier & 144 & Instance & 8 & 256 \\
    %     Plasticity & 64 & $L_2$ & Fourier & 144 & Instance & 8 & 128 \\
    %     Airfoil & 64 & $L_2$ & Eigensubspace & 64 & Layer & 6 & 64 \\
    %     Pipe & 4 & $L_2$ & Fourier & 144 & Layer & 8 & 96 \\
    % \bottomrule
    % \end{tabular}
    \begin{tabular}{c|c|ccccc}
    \toprule
        Test Case & Loss Fn & Basis & \#Basis & Normalization & \#Layers & \#Hiddens \\
    \midrule
        Darcy & $L_2 + 0.1 H_1$ & Chebyshev & 100 & Layer & 8 & 128 \\
        Navier Stokes & $L_2$ & Fourier & 144 & Instance & 8 & 256 \\
        Plasticity & $L_2$ & Fourier & 144 & Instance & 8 & 128 \\
        Airfoil & $L_2$ & Eigensubspace & 64 & Layer & 6 & 64 \\
        Pipe & $L_2$ & Fourier & 144 & Layer & 8 & 96 \\
    \bottomrule
    \end{tabular}
    \caption{The optimal configuration of SUPRA neural operator in Table \ref{tab:std-bench}.}
    \label{tab:optimal-configuration}
\end{table}

The following $L_2$ norm on domain $\Omega$ is used to compute our error metric:
\begin{equation}
    \| u \|_{2}^2 = \int_{\Omega} (u(x))^2 \mathrm{d}x \approx \frac{1}{M} \sum_{i\in \text{mesh}} u(x_i)^2,
\end{equation}
where $M$ is the number of sample points. We assume a structured or regular grid is $[0, 1]^2$. For the relative $L^2$ error used in accuracy comparison, coefficient $M$ is canceled:
\begin{equation}
    \text{Relative }L_2\text{ Error}(u, u^*) = \frac{\| u - u^*\|_2}{\|u^*\|_2} = \frac{\sqrt{\sum_{i\in \text{mesh}} (u(x_i) - u^*(x_i))^2}}{\sqrt{\sum_{i\in \text{mesh}} u^*(x_i)^2}}.
\end{equation}

%%%%%%%%%%%%%%%%%%%%%%%%%%%%%%%%%%%%%%%%%%%%%%%%%%%%%%%%%%%%%%%%%%%%%%%%%%%%%%%
%%%%%%%%%%%%%%%%%%%%%%%%%%%%%%%%%%%%%%%%%%%%%%%%%%%%%%%%%%%%%%%%%%%%%%%%%%%%%%%
\end{document}